\documentclass[10pt]{article}
\usepackage{amsthm}
\usepackage{amsmath,amssymb,amsfonts,amsbsy}
\usepackage[section]{algorithm}
\usepackage[noend]{algpseudocode}
\usepackage[affil-it]{authblk}
\usepackage[font=footnotesize,labelfont=bf]{caption}
\usepackage[left=1in,right=1in,top=1in,bottom=1in]{geometry}
\usepackage{graphicx}
\usepackage{hyperref}
\usepackage{mathrsfs}
\usepackage{physics}
\usepackage{titlesec}
\usepackage[dvipsnames]{xcolor}
\usepackage[normalem]{ulem}
\usepackage{url}
\usepackage{xspace}
\usepackage[nameinlink]{cleveref}

\hypersetup{
    colorlinks=true,
    linkcolor=black,
    citecolor=black,
    urlcolor=black,
    pdftitle={Active learning for data-driven reduced models of parametric differential systems with Bayesian operator inference},
    pdfauthor={Shane A. McQuarrie, Anirban Chaudhuri, Mengwu Guo}
}

\titleformat{\section}{\normalfont\large\bfseries}{\thesection}{1em}{}
\titleformat{\subsection}{\normalfont\normalsize\bfseries}{\thesubsection}{1em}{}

\theoremstyle{definition}
\newtheorem{remark}{Remark}[section]

\crefformat{equation}{#2(#1)#3}
\numberwithin{equation}{section}

\algnewcommand{\LineComment}[1]{\vspace{0.06125in}\newline\phantom{-----}\textcolor{gray}{\texttt{\# #1}}}
\algnewcommand{\LineCommentt}[1]{\vspace{0.06125in}\newline\phantom{---------}\textcolor{gray}{\texttt{\# #1}}}
\algrenewcomment[1]{\hfill\textcolor{gray}{\texttt{\# #1}}}
\algblock{Try}{EndTry}
\algcblock[Try]{Try}{Except}{EndTry}
\algcblockdefx{Try}{Except}{EndTry}[1]{\textbf{except} #1}

\algrenewtext{Try}{\textbf{try}}
\algtext*{EndTry}

\newcommand{\trp}{^{\mathsf{T}}}
\newcommand{\ddt}{\frac{\textrm{d}}{\textrm{d}t}}
\newcommand{\dxdt}[1]{\frac{\textrm{d}#1}{\textrm{d}t}}
\newcommand{\dt}{\,\textrm{d}t}
\DeclareMathOperator{\Var}{Var}

\title{\Large\textbf{Active learning for data-driven reduced models of parametric differential systems with Bayesian operator inference}}
\author[1]{Shane~A.~McQuarrie\thanks{Corresponding author. E-mail: \href{mailto:shanemcq@mathematics.byu.edu}{\texttt{shanemcq@mathematics.byu.edu}}.}}
\affil[1]{\normalsize Department of Mathematics, Brigham Young University}

\author[2]{Mengwu Guo}
\affil[2]{\normalsize Centre for Mathematical Sciences, Lund University}

\author[3]{Anirban Chaudhuri}
\affil[3]{\normalsize Oden Institute for Computational Engineering and Sciences, The University of Texas at Austin}

\date{}

\usepackage{fancyhdr}                      
\begin{document} % ============================================================

\thispagestyle{empty}

\maketitle

\vspace{-.5in}

\begin{abstract}
\noindent
This work develops an active learning framework to intelligently enrich data-driven reduced-order models (ROMs) of parametric dynamical systems, which can serve as the foundation of virtual assets in a digital twin. Data-driven ROMs are explainable, computationally efficient scientific machine learning models that aim to preserve the underlying physics of complex dynamical simulations. Since the quality of data-driven ROMs is sensitive to the quality of the limited training data, we seek to identify training parameters for which using the associated training data results in the best possible parametric ROM. Our approach uses the operator inference methodology, a regression-based strategy which can be tailored to particular parametric structure for a large class of problems. We establish a probabilistic version of parametric operator inference, casting the learning problem as a Bayesian linear regression. Prediction uncertainties stemming from the resulting probabilistic ROM solutions are used to design a sequential adaptive sampling scheme to select new training parameter vectors that promote ROM stability and accuracy globally in the parameter domain. We conduct numerical experiments for several nonlinear parametric systems of partial differential equations and compare the results to ROMs trained on random parameter samples. The results demonstrate that the proposed adaptive sampling strategy consistently yields more stable and accurate ROMs than random sampling does under the same computational budget.
 \end{abstract}

\vspace{2mm}

\section{Introduction}\label{sec:intro}

Numerical simulation of complex physical phenomena is a core enabling technology for digital twins, which are comprised of physical and virtual assets with a two-way flow of information: data from the physical asset is used to construct and/or calibrate the virtual asset (a numerical model), while numerical predictions from the virtual asset are used for control or decision-making for the physical asset \cite{naenasem2024digitaltwins}. To be viable for practical application, the virtual asset must be able to produce predictions rapidly and reliably; however, the underlying physics that are of interest for digital twin applications can typically only be accurately simulated using a large number of degrees of freedom, leading to computationally expensive numerical simulations. The explainability and computational efficiency of decisions made by the digital twin play a key role in safety-critical applications, making explainable artificial intelligence an essential ingredient~\cite{gunning2019darpa}. Model reduction techniques are one such explainable scientific machine learning technique that construct low-dimensional systems, called reduced-order models (ROMs), to serve as computationally inexpensive surrogates for a high-dimensional physics simulation~\cite{benner2015pmorsurvey,ghattas2021acta}. This paper introduces a technique for adaptively constructing ROMs to emulate systems with parametric dependence, that is, systems whose behavior varies with some set of parameters, usually representing physical properties. We focus on systems where the parametric dependence manifests in the operators defining the model, not merely in initial conditions or external inputs. Building parametric ROMs requires training data at different parameter combinations, which can incur significant training computational cost. Moreover, state-of-the-art parametric ROMs---especially those learned directly from data---are not typically equipped with automated methods for determining how many and which parameters to select for training data generation within a limited training budget. Our goal is to develop a framework for intelligently selecting training parameters in order to learn a high-quality ROM with as few training data as possible, in particular in the context of data-driven model reduction.

Classical model reduction techniques, including those that are designed for parametric systems, intrusively construct a ROM by compressing a large numerical model, called the full-order model (FOM), through direct projection~\cite{benner2015pmorsurvey}. On the other hand, non-intrusive data-driven model reduction techniques, such as dynamic mode decomposition (DMD)~\cite{brunton2015dmd,schmid2022dmd} and operator inference (OpInf)~\cite{ghattas2021acta,kramer2024survey,peherstorfer2016opinf}, construct a ROM by learning from available simulation data while imposing appropriate model structure without needing access to the implementation details of the FOM. The non-intrusive nature of these procedures eases the implementation burden and thus widens the applicability of model reduction. Since ROMs of parametric systems have application in large-scale optimization~\cite{ghattas2021acta,zahr2015progressive}, inverse problems~\cite{ghattas2021acta,lieberman2010inverse}, and uncertainty quantification~\cite{cicci2023gpromuq}, data-driven parametric model reduction is a particularly attractive prospect in the context of the digital twin framework \cite{kramer2026dtwinroms}. Examples of such techniques include the Loewner framework~\cite{ionita2014parametricloewner}, interpolation-based methods~\cite{franz2014manifoldinterp,zimmermann2021manifold}, parametric DMD~\cite{andreuzzi2023pDMD,duan2024pdmd,huhn2023pDMD,lin2023parametricGPDMD,sayadi2015pDMD}, parametric OpInf~\cite{farcas2023parametric,khodabakhshi2022daeopinf,mcquarrie2023parametric,vijaywargiya2025tensoropinf}, Gaussian process surrogate modeling~\cite{cicci2023gpromuq,guo2018gprom}, and deep learning~\cite{franco20203deeproms,fresca2021deeproms,ivagnes2023mlpipelineinverse}. Stability is a key issue for data-driven ROMs, and several recent works focus on embedding physical constraints or structure to promote or guarantee stability in OpInf ROMs. The work in \cite{zastrow2025} designs ROM structure to emulate precise coupling relationships between physical variables, while \cite{aretz2024nestedopinf,aretz2026nestedopinf} constrains learned operators to follow a hierarchical nested structure related to direct projection. Energy conservation for Hamiltonian systems is addressed in~\cite{gruber2025variational,sharma2022hamiltonian,vijaywargiya2025tensoropinf}, second-order systems for structural or Lagrangian dynamics are examined in~\cite{boef2024stablesparse,geng2024gradient,sharma2024lagrangianopinf}, and the works in~\cite{errico2026gasnitrom,goyal2025stablequadratic} obtain Lyapunov-based stability guarantees for quadratic ROMs. Other methods post-process the learned operators~\cite{sawant2023pireg} or state predictions at runtime~\cite{kim2025stateconstraints}; see also~\cite{kramer2021quadstability} for an analysis of stability of nonlinear polynomial models, including data-driven ROMs. Most of these stability-promoting methods are currently developed for non-parametric problems only. Our method leverages Tikhonov-regularized parametric OpInf~\cite{mcquarrie2023parametric}, not only because of its parametric structure, but also due to a natural connection to Bayesian inference~\cite{kaipio2005inverse,stuart2010inverse}.

Parametric data-driven models are only as good as the data they are trained on, and we can only obtain limited training data from expensive-to-evaluate complex systems due to training cost constraints. Therefore, what training data to obtain for optimal learning of the parametric ROMs within a given training budget is important to understand, since not all observations are equally useful. Active learning (or sequential optimal experimental design) is a way to select new training data given some existing knowledge about the current approximation, such as prediction uncertainty. In this work, the goal is to improve the global accuracy of the ROM by adaptively sampling the parameter space and, hence, selecting new training data. Some early work for active learning methods for global accuracy include maximizing mean squared error metrics~\cite{sacks1989design}, entropy~\cite{currin1988bayesian,sacks1989design,mackay1992bayesian}, and maximizing the expected reduction in predictive variance~\cite{cohn1996active} as acquisition functions for selecting the next training sample. The standard active learning literature uses Gaussian process regression (or Kriging)~\cite{rasmussen2006gaussian} as the surrogate model. There is limited work on active learning for adaptively updating parametric ROMs since they lack a measure of uncertainty in the predictions. Traditionally, greedy algorithms based on \emph{a posteriori} error estimators were used to generate snapshots for adaptive reduced basis construction~\cite{buffa2012,grepl2005posteriori,grepl2007nonlinearreducedbasis,quarteroni2011certified}; however, these methods are limited to the context of intrusive ROMs with certain structure that give rise to the error estimators. Finally, there are recent works on adaptive sampling for parametric ROMs~\cite{blais2025goal,zhuang2023activeMLrom} and for surrogate models fitted in the reduced input-output dimensions~\cite{guo2024active} that utilize additional surrogate models for error estimates to drive the parameter selection. In this work, we introduce an active learning method based on the prediction uncertainty from a probabilistic surrogate model, without the need for additional surrogates to provide ROM error estimates. Specifically, we propose a Bayesian parametric ROM equipped prediction uncertainty, which we use to design an active learning framework for adaptively selecting new training parameters.

The major contributions of this work are (i) a Bayesian parametric OpInf ROM method for affine-parametric systems that provides uncertainty in spatiotemporal predictions with respect to underlying model parameters and (ii) acquisition functions targeting stability and global accuracy for active learning in ROMs. We develop a probabilistic parametric ROM, building on our previous work in affine-parametric OpInf~\cite{mcquarrie2023parametric} and Bayesian OpInf~\cite{guo2022bayesopinf}. We derive closed form expressions for the posterior distribution of the affine reduced operators for the proposed Bayesian parametric OpInf. The Bayesian parametric OpInf is a probabilistic ROM that provides a measure of uncertainty in the predictions. The statistics obtained from the posterior ROM are used to formulate the proposed acquisition functions of the probability of ROM instability and the total prediction variance. Our iterative procedure does not directly enforce global stability for the entire parameter space at each step, but such stability follows naturally within in few iterations due to our choice of acquisition functions. We present numerical experiments on a heat equation with nonlinear reaction term and a two-dimensional viscous Burgers' problem for shock propagation. The results show that the proposed active learning methodology significantly results in satisfactory stability and accuracy at a fraction of the training computational cost when compared to random sampling of the parameter space.

The remainder of the paper is organized as follows.
\Cref{sec:opinf} describes the proposed Bayesian parametric ROM for affine-parametric systems.
\Cref{sec:implementation} uses the Bayesian parametric ROM to develop an active learning scheme for iteratively updating the training data for parametric OpInf ROMs. Implementation details and customization options are also discussed. The methodology is demonstrated on two problems in \Cref{sec:results}: a parametric diffusion-reaction equation with nonlinear reaction term, and a two-dimensional Burgers' equation with variable wave speed. Finally, \Cref{sec:conclusion} makes some concluding remarks and outlines avenues for future work.

\section{Bayesian parametric operator inference}\label{sec:opinf}

This section briefly reviews operator inference for affine-parametric systems as developed in \cite{mcquarrie2023parametric,yildiz2021shallow} in \Cref{sec:popinf}. Then we present the proposed Bayesian parametric operator inference method that provides prediction uncertainties in \Cref{sec:bayes}. The technique gives rise to a regression problem, which is endowed with a probabilistic interpretation to form a Bayesian linear regression.

\subsection{Non-intrusive reduced models for affine-parametric system}\label{sec:popinf}

Let $\vb*{\xi}\in\mathcal{P}\subset\mathbb{R}^{p}$ denote a vector of $p$ variables parameterizing a time-dependent partial differential equation (PDE). After introducing a spatial discretization, a finite-dimensional numerical model for the PDE may be written generically as
\begin{align}
    \label{eq:generic-fom}
    \ddt\vb{q}(t;\vb*{\xi})
    = {\vb{f}}({\vb{q}}(t), \vb*{\xi})\,,
    \quad
    \vb{q}(t_0;\vb*{\xi}) = \vb{q}_0(\vb*{\xi})\,,
    \quad
    t\in [t_0, t_f]\,,
\end{align}
where $\vb{q}(t;\vb*{\xi})\in\mathbb{R}^{N}$ is the time-dependent spatial discretization of the PDE state. This is the full-order model (FOM). The state depends implicitly on the parameter vector $\vb*{\xi}$ through the function $\vb{f}:\mathbb{R}^{N}\times\mathcal{P}\to\mathbb{R}^{N}$ which governs the dynamics, and through the initial condition $\vb{q}_0(\vb*{\xi})\in\mathbb{R}^{N}$. The function $\vb{f}$ may also depend on time-dependent system inputs, for example arising from nonstationary boundary conditions or source terms, but we neglect these for presentation clarity.

Many systems of interest can be written with the following parametric polynomial structure (we suppress the dependence of $\vb{q}(t;\vb*{\xi})$ on $t$ and $\vb*{\xi}$ going forward):
\begin{subequations}\label{eq:fom}
\begin{gather}
    \label{eq:polynomial-fom}
    \dxdt{\vb{q}}
    = \vb{c}(\vb*{\xi})
    + \vb{A}(\vb*{\xi})\vb{q}
    + \vb{H}(\vb*{\xi})[\vb{q}\otimes\vb{q}],
    \\
    \label{eq:fom-operators}
    \vb{c}(\vb*{\xi})\in\mathbb{R}^{N},
    \quad
    \vb{A}(\vb*{\xi})\in\mathbb{R}^{N\times N},
    \quad
    \vb{H}(\vb*{\xi})\in\mathbb{R}^{N\times N^2},
\end{gather}
where $\otimes$ denotes the Kronecker product \cite{vanLoan2000kronecker}. This work considers the class of problems where each of the ``operators'' in \cref{eq:fom-operators} depend affinely on the parameter vector $\vb*{\xi}$, meaning there exist
$\vb{c}^{(1)},\ldots,\vb{c}^{(n_c)}\in\mathbb{R}^{N}$,
$\vb{A}^{(1)},\ldots,\vb{A}^{(n_a)}\in\mathbb{R}^{N\times N}$, and
$\vb{H}^{(1)},\ldots,\vb{H}^{(n_h)}\in\mathbb{R}^{N\times N^2}$
such that
\begin{align}
    \label{eq:affine-fom}
    \begin{gathered}
    \vb{c}(\vb*{\xi})
    = \sum_{\ell=1}^{n_c}\theta_{c}^{(\ell)}\!(\vb*{\xi})\,\vb{c}^{(\ell)}\!,
    \qquad
    \vb{A}(\vb*{\xi})
    = \sum_{\ell=1}^{n_a}\theta_{a}^{(\ell)}\!(\vb*{\xi})\,\vb{A}^{\!(\ell)}\!,
    \\
    \vb{H}(\vb*{\xi})
    = \sum_{\ell=1}^{n_h}\theta_{h}^{(\ell)}\!(\vb*{\xi})\,\vb{H}^{(\ell)}\!,
    \end{gathered}
\end{align}
\end{subequations}
where each $\theta_x^{(\ell)}:\mathcal{P}\to\mathbb{R}$ is a scalar-valued function of the parameter vector. In many cases, including the numerical examples presented later, the number of terms $n_c$, $n_a$, and $n_h$ in each expansion is small.

State-polynomial, affine-parametric structure such as in \cref{eq:fom} is advantageous for projection-based model reduction because linear dimensionality reduction preserves this structure. Given a matrix $\vb{V}\in\mathbb{R}^{N\times r}$ with orthonormal columns, making the low-dimensional approximation $\vb{q} \approx \vb{V}\hat{\vb{q}}$ for some $\hat{\vb{q}} = \hat{\vb{q}}(t;\vb*{\xi})\in\mathbb{R}^{r}$ leads to the low-dimensional evolution equations $\ddt\hat{\vb{q}} = \vb{V}\trp\vb{f}(\vb{V}\hat{\vb{q}},\vb*{\xi})$. For \cref{eq:fom}, this can be written as
\begin{subequations}\label{eq:rom}
\begin{gather}
    \label{eq:polynomial-rom}
    \dxdt{\hat{\vb{q}}}
    = \hat{\vb{c}}(\vb*{\xi})
    + \hat{\vb{A}}(\vb*{\xi})\hat{\vb{q}}
    + \hat{\vb{H}}(\vb*{\xi})[\hat{\vb{q}}\otimes\hat{\vb{q}}],
    \\
    \label{eq:rom-operators}
    \hat{\vb{c}}(\vb*{\xi})\in\mathbb{R}^{r},
    \quad
    \hat{\vb{A}}(\vb*{\xi})\in\mathbb{R}^{r\times r},
    \quad
    \hat{\vb{H}}(\vb*{\xi})\in\mathbb{R}^{r\times r^2}.
\end{gather}
Critically, the affine-parametric structure \cref{eq:affine-fom} is preserved in the reduced system matrices: there exist
$\hat{\vb{c}}^{(1)},\ldots,\hat{\vb{c}}^{(n_c)}\in\mathbb{R}^{r}$,
$\hat{\vb{A}}^{(1)},\ldots,\hat{\vb{A}}^{(n_a)}\in\mathbb{R}^{r\times r}$, and
$\hat{\vb{H}}^{(1)},\ldots,\hat{\vb{H}}^{(n_h)}\in\mathbb{R}^{r\times r^2}$
with
\begin{align}
    \label{eq:affine-rom}
    \begin{gathered}
    \hat{\vb{c}}(\vb*{\xi})
    = \sum_{\ell=1}^{n_c}\theta_{c}^{(\ell)}\!(\vb*{\xi})\,\hat{\vb{c}}^{(\ell)}\!,
    \qquad
    \hat{\vb{A}}(\vb*{\xi})
    = \sum_{\ell=1}^{n_a}\theta_{a}^{(\ell)}\!(\vb*{\xi})\,\hat{\vb{A}}^{\!(\ell)}\!,
    \\
    \hat{\vb{H}}(\vb*{\xi})
    = \sum_{\ell=1}^{n_h}\theta_{h}^{(\ell)}\!(\vb*{\xi})\,\hat{\vb{H}}^{(\ell)}\!,
    \end{gathered}
\end{align}
\end{subequations}
where each expansion has the same number of terms $n_x$ and the same coefficient functions $\theta_x^{(\ell)}$ as in \cref{eq:affine-fom}. This means that if the polynomial form and the affine-parametric structure of a high-dimensional model are known, we can stipulate the same structure in a low-dimensional space to design an effective ROM. The main advantage is that if $r \ll N$, solving \cref{eq:rom} is much less computationally expensive than solving \cref{eq:fom}.

\begin{remark}[Dimensionality reduction]
\label{remark:reduction}
The linearity of the approximation $\vb{q} \approx \vb{V}\hat{\vb{q}}$ is the reason that the structure of \cref{eq:fom} is preserved in \cref{eq:rom}. If a shifted approximation is made, i.e., $\vb{q} \approx \vb{V}\hat{\vb{q}} + \bar{\vb{q}}$ for a fixed nonzero reference vector $\bar{\vb{q}}\in\mathbb{R}^{N}$, additional terms arise in the reduced dynamics, but the new structure is straightforward to derive. The situation is further complicated by partially nonlinear approximations, such as quadratic manifolds~\cite{barnett2022quadmanifold,geelen2023quadmanifold,jain2017quadratic,schwerdtner2024greedy}, but the model form can usually still be written out explicitly. However, when using a fully nonlinear approximation, for example with autoencoders~\cite{champion2019autoencodersindy,chen2023crom,lee2020autoencoder}, it is  no longer straightforward to write down the expected parametric structure of the reduced dynamics.
\end{remark}

Operator inference (OpInf) \cite{ghattas2021acta,kramer2024survey,peherstorfer2016opinf} is a non-intrusive strategy for learning ROMs with polynomial structure. In the original work \cite{peherstorfer2016opinf}, parametric models of the form \cref{eq:polynomial-fom}--\cref{eq:fom-operators} are constructed by learning nonparametric models for each of several training parameter instances, then the learned models are interpolated in parameter space to extrapolate to new parameter instances. Later, \cite{mcquarrie2023parametric,yildiz2021shallow} extended the methodology to the case of affine-parametric structure as in~\cref{eq:affine-fom}. Affine-parametric OpInf (pOpInf) learns ROMs that can be written generally as
\begin{align}
    \label{eq:opinf-rom}
    \dxdt{\hat{\vb{q}}}
    = \hat{\vb{O}}\vb{d}(\hat{\vb{q}}; \vb*{\xi})
    \,,\quad
    \hat{\vb{q}}(t_0) = \hat{\vb{q}}_0\,, \quad t\in [t_0, t_f]\,,
\end{align}
where the operator matrix $\hat{\vb{O}}\in\mathbb{R}^{r\times d(r)}$ contains the (parameter-independent) vectors and matrices defining the ROM and $\vb{d}(\hat{\vb{q}};\vb*{\xi})\in\mathbb{R}^{d(r)}$ encodes the polynomial and parametric structure of the system. In the case of \cref{eq:rom}, we have
\begin{subequations}
\begin{align}
    \label{eq:quadratic-operator-matrix}
    \hat{\vb{O}}
    &= \left[\begin{array}{ccc|ccc|ccc}
        \hat{\vb{c}}^{(1)} & \cdots & \hat{\vb{c}}^{(n_c)} &
        \hat{\vb{A}}^{(1)} & \cdots & \hat{\vb{A}}^{(n_a)} &
        \hat{\vb{H}}^{(1)} & \cdots & \hat{\vb{H}}^{(n_h)}
\end{array}\right],
    \\
    \vb{d}(\hat{\vb{q}}, \vb{u}; \vb*{\xi})
    &= \left[\begin{array}{cccccc}
        \theta_{c}^{(1)}\!(\vb*{\xi})
        & \cdots &
        \theta_{c}^{(n_c)}\!(\vb*{\xi})
        &
        \theta_{a}^{(1)}\!(\vb*{\xi})\hat{\vb{q}}\trp
        & \cdots &
\theta_{h}^{(n_h)}\!(\vb*{\xi})[\hat{\vb{q}}\otimes\hat{\vb{q}}]\trp
\end{array}\right]\trp,
\end{align}
\end{subequations}
with $d(r) = n_c + r n_a + r^2 n_h$. The vector-valued function $\vb{d}$ is known \emph{a priori} because the structure of the ROM \cref{eq:opinf-rom} is stipulated to match the structure of a classical projection-based ROM, which, in the case being examined, inherits the structure of the original model. The operator matrix $\hat{\vb{O}}$, on the other hand, is not known \emph{a priori} in cases where the high-dimensional operators (e.g., $\vb{A}^{(1)},\ldots,\vb{A}^{(n_a)})$ are unknown or inaccessible. The goal is to learn an appropriate $\hat{\vb{O}}$ from data, as we now explain.

Consider a limited number of training parameter instances $\vb*{\xi}_{1},\ldots,\vb*{\xi}_{n_p}$ for which solution data are available from the FOM \cref{eq:fom}. Let $\{\vb{q}_{ij}\}_{j=1}^{n_t}$ be snapshots of the state solution of~\cref{eq:fom} at $n_t$ time instances $t_1,\ldots,t_{n_t}$ with parameter vector $\vb*{\xi} = \vb*{\xi}_{i}$. These training states are compressed to produce reduced snapshots $\hat{\vb{q}}_{ij} = \vb{V}\trp\vb{q}_{ij}$ and then used to estimate time derivatives $\dot{\hat{\vb{q}}}_{ij} \approx \ddt\vb{q}(t;\vb*{\xi}_{i})|_{t=t_j}$, for example with finite differences. The operator matrix is selected by solving the regression problem
\begin{subequations}\label{eq:popinf-both}
\begin{align}
    \label{eq:popinf-regression}
    \min_{\hat{\vb{O}}}
    \sum_{i=1}^{n_p}\sum_{j=1}^{n_t}\left\|
        \hat{\vb{O}}\vb{d}(\hat{\vb{q}}_{ij};\vb*{\xi}_{i})
        - \dot{\hat{\vb{q}}}_{ij}
    \right\|_{2}^{2},
\end{align}
which minimizes the residual of \cref{eq:opinf-rom} with respect to available observations. Letting $\vb{d}_{i,j} = \vb{d}(\hat{\vb{q}}_{ij},;\vb*{\xi}_{i})$ and defining the data matrices
\begin{align}
    \label{eq:popinf-ddtmatrix}
    \dot{\hat{\vb{Q}}}
    &= [~\dot{\hat{\vb{q}}}_{1,1}~~\cdots~~\dot{\hat{\vb{q}}}_{1,n_t}~~\dot{\hat{\vb{q}}}_{2,1}~~\cdots~~\dot{\hat{\vb{q}}}_{n_p,n_t}~]
    \in\mathbb{R}^{r\times n_p n_t},
    \\
    \label{eq:popinf-datamatrix}
    \vb{D}
    &= [~\vb{d}_{1,1}~~\cdots~~\vb{d}_{1,n_t}~~\vb{d}_{2,1}~~\cdots~~\vb{d}_{n_p,n_t}~]\trp
    \in\mathbb{R}^{n_p n_t \times d(r)},
\end{align}
we can write \cref{eq:popinf-regression} in the standard linear least-squares form
\begin{align}
    \label{eq:popinf-matricized}
    \min_{\hat{\vb{O}}}\left\|
        \vb{D}\hat{\vb{O}}\trp
        - \dot{\hat{\vb{Q}}}\trp
    \right\|_{F}^{2}\,.
\end{align}
Furthermore, this regression decouples into $r$ independent least-squares problems. Letting $\hat{\vb{o}}_{k}\in\mathbb{R}^{d(r)}$ and $\vb{z}_{k}\in\mathbb{R}^{n_p n_t}$ denote the $k$-th rows of $\hat{\vb{O}}$ and $\dot{\hat{\vb{Q}}}$, respectively, \cref{eq:popinf-matricized} can be written as
\begin{align}
    \label{eq:popinf-regression-decoupled}
    \min_{\hat{\vb{o}}_k}\left\|
        \vb{D}\hat{\vb{o}}_k
        - \vb{z}_k
        \right\|_{2}^{2}\,,
    \quad
    k = 1,\ldots,r,
    \qquad\qquad
    \begin{aligned}
    \hat{\vb{O}}
    &= [~\hat{\vb{o}}_{1}~~\cdots~~\hat{\vb{o}}_{r}~]\trp,
    \\
    \dot{\hat{\vb{Q}}}
    &= [~\vb{z}_{1}~~\cdots~~\vb{z}_{r}~]\trp,
    \end{aligned}
\end{align}
\end{subequations}
where each subproblem can be solved independent of the others.

Data-driven ROMs \cref{eq:opinf-rom} generated through the pOpInf regression \cref{eq:popinf-both} are not guaranteed to be stable. One standard method for promoting model stability is to add a regularization term to the regression, replacing~\cref{eq:popinf-regression-decoupled} with, for instance, a Tikhonov-regularized (ridge) regression,
\begin{align}
    \label{eq:popinf-regression-regularized}
    \min_{\hat{\vb{o}}_k}\left\{
        \left\|
            \vb{D}\hat{\vb{o}}_k
            - \vb{z}_k
        \right\|_{2}^{2}
        + \left\|
            \vb*{\Gamma}_{\!k}\hat{\vb{o}}_{k}
        \right\|_{2}^{2}
    \right\}\,,
    \qquad
    k = 1, \ldots, r,
\end{align}
where $\vb{\Gamma}_{\!k} \in \mathbb{R}^{d(r)\times d(r)}$. The regularization matrix $\vb{\Gamma}_{\!k}$ is often parameterized with a small number of hyperparameters and designed so that like terms are penalized equally \cite{mcquarrie2021combustion,qian2022pdes,sawant2023pireg}. Alternative strategies for promoting or even enforcing stability typically involve constraining the optimization problem \cref{eq:popinf-both}, for example to preserve certain structure \cite{aretz2024nestedopinf,boef2024stablesparse,errico2026gasnitrom,geng2024gradient,goyal2025stablequadratic,gruber2025variational,sharma2022hamiltonian,vijaywargiya2025tensoropinf,zastrow2025}. The lack of constraints in \cref{eq:popinf-regression-regularized} and our choice of Tikhonov regularization does not guarantee stability, but it admits a critical connection to Bayesian inference~\cite{kaipio2005inverse,stuart2010inverse} that enables a framework for constructing probabilistic ROMs.

\subsection{Bayesian framework for affine-parametric operator inference}\label{sec:bayes}

We now present the Bayesian parametric OpInf method that can account for underlying model errors to provide prediction uncertainties. The reduced snapshot time derivatives $\dot{\hat{\vb{q}}}_{ij}$ are typically estimated from the reduced snapshots $\hat{\vb{q}}_{ij}$, incurring errors to which the regression problems \cref{eq:popinf-both} and \cref{eq:popinf-regression-regularized} are sensitive. Modeling the estimation error as uncorrelated Gaussian noise, we obtain the Bayesian linear regression
\begin{align}
    \label{eq:bayes-regression}
    \vb{z}_{k}
    = \vb{D}\hat{\vb{o}}_{k} + \vb*{\varepsilon}_{k},
    \qquad
    p(\vb*{\varepsilon}_{k})
    = \mathcal{N}(\vb*{\varepsilon}_{k}\mid\vb{0}_{n_p n_t}, \sigma_k^2\vb{I}),
    \qquad
    k = 1, \ldots, r,
\end{align}
in which $\vb{0}_{n_p n_t}\in\mathbb{R}^{n_p n_t}$ denotes the zero vector and $\vb{I}$ is the $n_p n_t \times n_p n_t$ identity matrix. Each $\sigma_k^2 > 0$ describes the noise variance in the regression for $\hat{\vb{o}}_{k}$, the $k$-th row of the operator matrix $\hat{\vb{O}}$. The vector $\boldsymbol{\varepsilon}_k\in\mathbb{R}^{n_p n_t}$ corresponds to noise over the $n_t$ time instances for each of the $n_p$ training parameter samples. The noise is not correlated in time or across parameter values. The formulation builds on our previous work in \cite{guo2022bayesopinf}. The noise model in \eqref{eq:bayes-regression} assumes that the errors in time-derivatives are uncorrelated across time and parameter values. While we estimate the time-derivatives non-sequentially from the data, finite difference approximations generally share neighboring observations that introduce some correlation in the time-derivative errors. However, in this work, we adopt the independent Gaussian noise model as a simplifying approximation (i.e., without defining a specific temporal correlation structure in the trajectories), enabling computationally tractable Bayesian inference.

We immediately have a Gaussian likelihood function for $\hat{\vb{o}}_{k}$ given data in $\vb{D}$:
\begin{align}
    \label{eq:bayes-likelihood}
    \begin{aligned}
    \pi_\text{like}(\hat{\vb{o}}_k)
    = p(\vb{z}_{k}\mid\vb{D},\hat{\vb{o}}_k,\sigma_{k}^{2})
    &= \mathcal{N}(\vb{z}_{k}\mid\vb{D}\hat{\vb{o}}_k, \sigma_{k}^{2}\vb{I})
\\ &
    \propto \exp\left(-\frac{1}{2\sigma_k^{2}}\|\vb{D}\hat{\vb{o}}_{k} - \vb{z}_{k}\|_2^{2}\right).
    \end{aligned}
\end{align}
To facilitate a Bayesian inference, we impose a Gaussian prior on the operator vector, namely
\begin{align}
    \label{eq:bayes-prior}
    \begin{aligned}
    \pi_\text{prior}(\hat{\vb{o}}_k)
    &= p(\hat{\vb{o}}_{k}\mid\vb*{\gamma}_k,\sigma_{k}^{2})
    \\
    &= \mathcal{N}(\hat{\vb{o}}_{k}\mid\vb{0}_{d(r)}, \sigma_{k}^{2}\operatorname{diag}(\vb*{\gamma}_k)^{-1})
    \\
    &\propto \exp\left(-\frac{1}{2\sigma_{k}^{2}}\hat{\vb{o}}_{k}\trp\operatorname{diag}(\vb*{\gamma}_{\!k})\hat{\vb{o}}_{k}\right)
    = \exp\left(-\frac{1}{2\sigma_{k}^{2}}\big\|\operatorname{diag}(\vb*{\gamma}_{\!k})^{1/2}\hat{\vb{o}}_{k}\big\|_{2}^{2}\right)
    \end{aligned}
\end{align}
where $\vb*{\gamma}_1,\ldots,\vb*{\gamma}_{r}\in\mathbb{R}^{d(r)}$, $\vb{0}_{d(r)}\in\mathbb{R}^{d(r)}$ is the zero vector, and $\sigma_{k}^{2}\operatorname{diag}(\vb*{\gamma}_{\!k})^{-1}$ is the prior covariance for $\hat{\vb{o}}_{k}$, $k=1,\ldots,r$. The scalar likelihood variance $\sigma_{k}^{2}$ is included in the prior for convenience in matching to terms in \cref{eq:bayes-likelihood}. Equipping the prior with a diagonal covariance implies that the entries of the operator vector $\hat{\vb{o}}_k$ are uncorrelated, and the zero prior mean disincentives the entries of $\hat{\vb{o}}_k$ from becoming too large. Note that this also follows from the explicit connection between Tikhonov regularization and Bayesian linear regression~\cite{kaipio2005inverse,stuart2010inverse}.

The goal now is to establish a probability distribution for $\hat{\vb{o}}_{k}$ given training data for $\vb{D}$ and $\vb{z}_{k}$. Applying Bayes' rule, we obtain
\begin{align}
    \label{eq:bayes-posterior}
    \begin{aligned}
    \pi_\text{post}(\hat{\vb{o}}_k)
&\propto
    \pi_\text{like}(\hat{\vb{o}}_k)\pi_\text{prior}(\hat{\vb{o}}_k)
    \\
    &\propto \exp\left(-\frac{1}{2\sigma_k^{2}}\big\|\vb{D}\hat{\vb{o}}_{k} - \vb{z}_{k}\big\|^{2} -\frac{1}{2\sigma_{k}^{2}}\big\|\operatorname{diag}(\vb*{\gamma}_{\!k})^{1/2}\hat{\vb{o}}_{k}\big\|_{2}^{2}\right)
    \\
    &\propto \exp\left(-\frac{1}{2}(\hat{\vb{o}}_{k} - \vb*{\mu}_{k})\trp\vb*{\Sigma}_{k}^{-1}(\hat{\vb{o}}_{k} - \vb*{\mu}_{k})\right)\,,
\end{aligned}
\end{align}
that is,
$
    \pi_\text{post}(\hat{\vb{o}}_{k})
    \propto \mathcal{N}(\hat{\vb{o}}_k \mid \vb*{\mu}_k,\vb*{\Sigma}_k)\,,
$
with posterior mean and covariance
\begin{subequations}
\label{eq:bayes-posterior-parameters}
\begin{align}
    \vb*{\mu}_{k}
    &= \left(\vb{D}\trp\vb{D} + \operatorname{diag}(\vb*{\gamma}_k)\right)^{-1}\vb{D}\trp\vb{z}_{k}
= \underset{\vb*{\mu}}{\operatorname{argmin}}\Big\{
        \|\vb{D}\vb*{\mu} - \vb{z}_{k}\|_{2}^{2}
        + \|\vb*{\Gamma}_{\!k}\vb*{\mu}\|_2^2
    \Big\},
    \\
    \vb*{\Sigma}_{k}
    &= \sigma_{k}^{2}\left(\vb{D}\trp\vb{D} + \operatorname{diag}(\vb*{\gamma}_{k})\right)^{-1}
    = \sigma_{k}^{2}\left(\vb{D}\trp\vb{D} + \vb*{\Gamma}_{\!k}\trp\vb*{\Gamma}_{\!k}\right)^{-1},
\end{align}
wherein $\vb*{\Gamma}_{\!k} = \operatorname{diag}(\vb*{\gamma}_{\!k})^{1/2}$. A maximum marginal likelihood argument yields optimal variance values
\begin{align}
    \sigma_{k}^{2}
    = \frac{1}{n_p n_t}\Big(
        \|\vb{D}\vb*{\mu}_{k} - \vb{z}_{k}\|_2^2
        + \|\vb*{\Gamma}_{\!k}\vb*{\mu}_{k}\|_2^2
    \Big),
\end{align}
\end{subequations}
see \cite{guo2022bayesopinf} for the derivation of this estimator, noting that $\vb*{\mu}_{k}$ solves the regression \cref{eq:popinf-regression-regularized} with regularization
matrix $\vb*{\Gamma}_{\!k}$. It is also worth pointing out that, though the Bayesian framework in this subsection shares the same mathematical derivation as that in \cite{guo2022bayesopinf}, the assembly of $\hat{\vb{O}}$ takes the affine parameter-dependency into account, and the data matrices $\vb{D}$ and $\dot{\hat{\vb{Q}}}$ include solution data evaluated at different parameter vectors.

The definitions in \cref{eq:bayes-posterior-parameters} establish a probability distribution for the operator matrix $\vb{\hat{O}}$ in the parametric model~\cref{eq:opinf-rom}. To generate a single sample from this distribution, draw $\hat{\vb{o}}_{k}^{(1)}$ from $\mathcal{N}(\vb*{\mu}_{k},\vb*{\Sigma}_{k})$ for each $k = 1,\ldots,r$ and stack the results to form
$\hat{\vb{O}}^{(1)} = [~\hat{\vb{o}}_{1}^{(1)}~~\cdots~~\hat{\vb{o}}_{r}^{(1)}~]\trp$. Note that this is a computationally inexpensive task. Since the vector-valued function $\vb{d}(\hat{\vb{q}};\vb*{\xi})$ is pre-specified, fixing $\hat{\vb{O}} = \hat{\vb{O}}^{(1)}$ fully defines~\cref{eq:opinf-rom} as a deterministic ROM which can be solved for any parameter vector $\vb*{\xi}\in\mathcal{P}$. On the other hand, considering $\hat{\vb{O}}$ as a matrix-valued random variable defines \cref{eq:opinf-rom} as a probabilistic ROM whose state solution, which we denote $\hat{\vb{q}}_\text{rom}(t;\vb*{\xi})$, is a stochastic process. The moments of the state solution can be estimated efficiently with Monte Carlo sampling and will be used shortly to drive our active learning procedure.

\begin{algorithm}[t]
\begin{algorithmic}[1]
\Procedure{pBayesOpInf}{
    \newline\phantom{---}
    Training parameter vectors $\vb*{\xi}_{1},\ldots,\vb*{\xi}_{n_p}\in\mathcal{P}$,
    \newline\phantom{---}
    Compressed states $\{\hat{\vb{q}}_{ij}\}_{j=1}^{n_t}\subset\mathbb{R}^{r}$ corresponding to $\vb*{\xi}_{i}$ for $i=1,\ldots,n_p$,
    \newline\phantom{---}
    Parametric ROM structure function $\vb{d} = \vb{d}(\hat{\vb{q}};\vb*{\xi})$,
    \newline\phantom{---}
    Number of posterior draws $n_d > 0$
    \newline}
\LineComment{Set up the OpInf regression.}
    \State $\dot{\hat{\vb{Q}}} = [~\vb{z}_{1}~~\cdots~~\vb{z}_{r}~]\trp \gets$~\texttt{ddt(}$\hat{\vb{q}}_{ij}~\forall i,j$\texttt{)}
        \Comment{Time derivatives \cref{eq:popinf-ddtmatrix}.}
    \State $\vb{D} \gets~$\texttt{datamatrix(}$\vb{d}(\hat{\vb{q}}_{ij};\vb*{\xi}_{i})~\forall i,j$\texttt{)}
        \Comment{Data matrix \cref{eq:popinf-datamatrix}.}
    \State $\vb*{\Gamma} \gets$~\texttt{regselect(}$\hat{\vb{q}}_{ij},\vb*{\xi}_{ij}~\forall i,j$\texttt{)}
        \Comment{Select regularization matrix.}\label{step:regselect}\LineComment{Construct the operator matrix distribution.}
    \For{$k = 1,\ldots, r$}
        \State $\vb*{\mu}_k \gets \operatorname{argmin}_{\vb*{\mu}}\big\{
        \|\vb{D}\vb*{\mu} - \vb{z}_{k}\|_{2}^{2}
        + \|\vb*{\Gamma}\vb*{\mu}\|_2^2\big\}$
            \Comment{Posterior mean for $\hat{\vb{o}}_{k}$.}
        \State $\sigma_{k}^{2} \gets \frac{1}{n_p n_t}\Big(
        \|\vb{D}\vb*{\mu}_{k} - \vb{z}_{k}\|_2^2
        + \|\vb*{\Gamma\mu}_{k}\|_2^2
    \Big)$
        \State $\vb*{\Sigma}_{k} \gets \sigma_{k}^{2}\big(\vb{D}\trp\vb{D} + \vb*{\Gamma}\trp\vb*{\Gamma}\big)^{-1}$
            \Comment{Posterior covariance for $\hat{\vb{o}}_{k}$.}
    \EndFor
    \LineComment{Draw from the operator matrix distribution.}
    \For{$\ell=1,\ldots,n_d$}
        \For{$k = 1,\ldots, r$}
            \State $\hat{\vb{o}}_{k}^{(\ell)}\gets$~draw from $\mathcal{N}(\vb*{\mu}_{k},\vb*{\Sigma}_{k})$
        \EndFor
        \State $\hat{\vb{O}}^{(\ell)} \gets [~\hat{\vb{o}}_{1}^{(\ell)}~~\cdots~~\hat{\vb{o}}_{r}^{(\ell)}~]\trp$
    \EndFor
    \State \textbf{return} $\hat{\vb{O}}^{(1)},\ldots,\hat{\vb{O}}^{(n_d)}$
\EndProcedure
\end{algorithmic}
\caption{Bayesian affine-parametric operator inference.}
\label{alg:pBayesOpInf}
\end{algorithm}

The only outstanding information needed to construct a probabilistic ROM as described above is the prior hyperparameters $\vb*{\gamma}_{1},\ldots,\vb*{\gamma}_{r}$ or, equivalently, the diagonal regularization matrices $\vb*{\Gamma}_{1},\ldots,\vb*{\Gamma}_{r}$. To select these, we adapt the approaches from \cite{guo2022bayesopinf,mcquarrie2025gpbayesopinf,mcquarrie2021combustion,mcquarrie2023parametric}, which set $\vb*{\gamma}_{1} = \vb*{\gamma}_{2} = \cdots = \vb*{\gamma}_{r}$ and parameterize $\vb{\Gamma} = \operatorname{diag}(\vb*{\gamma}_{k})$ so that nonlinear terms are regularized separately from remaining terms. In the quadratic model \cref{eq:rom} with operator matrix \cref{eq:quadratic-operator-matrix}, this results in the total regularization term
\begin{align}
    \sum_{k=1}^{r}\big\|\vb*{\Gamma}\hat{\vb{o}}_{k}\big\|_{2}^{2}
    = \lambda_{1}\left(
        \sum_{\ell=1}^{n_c}\big\|\hat{\vb{c}}_{\ell}\big\|_{2}^{2}
        + \sum_{\ell=1}^{n_a}\big\|\hat{\vb{A}}_{\ell}\big\|_{F}^{2}
    \right)
    + \lambda_{2}\sum_{\ell=1}^{n_h}\big\|\hat{\vb{H}}_{\ell}\big\|_{F}^{2}.
\end{align}
The scalars $\lambda_{1},\lambda_{2}\ge 0$ are selected through a grid search that prioritizes stability. Given a pair of candidate regularization values $(\lambda_1,\lambda_2)$, we construct the operator matrix posterior \cref{eq:bayes-posterior}--\cref{eq:bayes-posterior-parameters} and, for each training parameter vector $\vb*{\xi}_{i}$, sample the posterior $n_d > 0$ times and solve the corresponding deterministic ROMs at that parameter vector. If any of the $n_d$ draws result in an unstable ROM, the candidate pair is discarded. In this context, stability refers to numerical stability over the time domain of interest, measured by whether the ROM state exceeds a large predetermined value. If all ROM solutions are stable in this sense, they are averaged over the draws and compared to the training snapshots. More precisely, we draw operator matrices $\hat{\vb{O}}^{(1)},\ldots,\hat{\vb{O}}^{(n_d)}$ and estimate the process mean with the sample mean
\begin{align}
    \mathbb{E}\big[\hat{\vb{q}}_\textrm{rom}(t;\vb*{\xi}_{i})\big]
    \approx
    \overline{\hat{\vb{q}}_\textrm{rom}(t;\vb*{\xi}_{i})}
    = \frac{1}{n_d}\sum_{\ell=1}^{n_d}\hat{\vb{q}}(t;\vb*{\xi}_{i};\hat{\vb{O}}^{(\ell)}),
\end{align}
where $\hat{\vb{q}}(t;\vb*{\xi}_{i};\hat{\vb{O}}^{(\ell)})$ is the solution to \cref{eq:opinf-rom} with $\hat{\vb{O}} = \hat{\vb{O}}^{(\ell)}$ and $\vb*{\xi} = \vb*{\xi}_{i}$. Then, we compute the average relative training reconstruction error of the sample mean,
\begin{align}
    \label{eq:training-error}
    \mathcal{E}_\textrm{train} =
    \frac{1}{n_p n_t}\sum_{i=1}^{n_p}\sum_{j=1}^{n_t}\frac{\left\|
        \hat{\vb{q}}_{i,j} - \overline{\hat{\vb{q}}_\textrm{rom}(t_{j};\vb*{\xi}_{i})}
    \right\|_{2}^{2}}{
    \left\|\hat{\vb{q}}_{i,j}\right\|_{2}^{2}
    }.
\end{align}
The regularization hyperparameters $(\lambda_1,\lambda_2)$ that result in the lowest training error \cref{eq:training-error} without any unstable draws are selected to define the final probabilistic ROM. This procedure can be carried out through a grid search and fine tuned with derivative-free optimization techniques. The approach balances stability with accuracy at the training parameter vectors: though there is no guarantee that all possible draws of the probabilistic ROM will be stable at the training parameter vectors, we observe empirically in the numerical experiments that if a modest $n_d = 20$ draws are stable, additional draws tend to be stable. However, stability at the training parameter vectors does not guarantee stability at non-training parameter vectors.

\Cref{alg:pBayesOpInf} details the procedure for constructing and sampling from the posterior operator matrix distribution \cref{eq:bayes-posterior}. The \texttt{regselect()} procedure in \cref{step:regselect} represents the regularization selection process described in the previous paragraph.

\section{Active learning for Bayesian parametric OpInf ROMs}\label{sec:implementation}

\begin{algorithm}[t]
\begin{algorithmic}[1]
\Procedure{BayesOpInfAcquisition}{
    \newline\phantom{---}
    Test parameter vector $\tilde{\vb*{\xi}}$,
    \newline\phantom{---}
    Operator matrix realizations $\hat{\vb{O}}^{(1)},\ldots,\hat{\vb{O}}^{(n_d)}\in\mathbb{R}^{r\times d(r)}$
\newline\phantom{---}
Parametric ROM structure function $\vb{d} = \vb{d}(\hat{\vb{q}};\vb*{\xi})$,
\newline}
\State $n_\alpha \gets 0$
        \Comment{Counter for number of unstable draws.}
    \State $S \gets \{\}$
        \Comment{Set for the indices of stable draws.}
    \For{$\ell = 1, \ldots, n_d$}
        \Try
            \State $\tilde{\vb{Q}}^{(\ell)} \gets$~integrate $\dxdt{\hat{\vb{q}}} = \hat{\vb{O}}^{(\ell)}\vb{d}(\hat{\vb{q}}(t);\tilde{\vb*{\xi}})$
            \State Add $\ell$ to the indexing set $S$
        \Except{time integration is numerically unstable}
            \State $n_\alpha \gets n_\alpha + 1$
        \EndTry
    \EndFor
    \State $\alpha \gets n_\alpha \,/\, n_d$
        \Comment{Probability of ROM instability.}
    \If{$\alpha = 100\%$ (all $n_d$ time integrations unstable)}
        \State $\omega \gets $ \texttt{None}
            \Comment{No accuracy acquisition values.}
    \Else
        \State $\omega \gets \hat{\varphi}(\tilde{\vb*{\xi}}; \{\tilde{\vb{Q}}^{(\ell)}\}_{\ell\in S})$
        \Comment{Global accuracy acquisition value~\cref{eq:acquisition-practical}.}
    \EndIf
    \State \textbf{return} $\alpha, \omega$
\EndProcedure
\end{algorithmic}
\caption{Posterior sampling and acquisition evaluation with parametric BayesOpInf.}
\label{alg:BayesOpInfAcquisition}
\end{algorithm}

The performance quality of data-driven models, including operator inference ROMs, depends strongly on the quality of the training data. In the model reduction setting, acquiring new training data amounts to either solving an expensive FOM or performing a physical experiment and collecting observations. The goal of this section is to present a novel active learning method that can use Bayesian pOpInf to iteratively determine new points in the parameter domain such that new training data corresponding to these points are optimal in some sense, allowing us to learn accurate pOpInf ROMs with as little training data as is practical and provide prediction uncertainties.

\subsection{Acquisition functions for new training parameters}

Given a probabilistic pOpInf ROM constructed through \Cref{alg:pBayesOpInf}, our goal is to choose a new parameter vector, not yet used for training, at which to solve the FOM and supplement the training set. There are two main considerations to use as selection criteria: (i) ROM stability and (ii) global accuracy of predictions. The regularization selection procedure promotes ROM stability at the training parameter vectors. However, we do not have guarantees for ROM performance in terms of accuracy, uncertainty, or stability at non-training parameter vectors, nor do we know beforehand how adding a particular parameter vector to the training set might improve ROM performance throughout the parameter domain. We choose parameter vectors based on two proposed active learning acquisition functions that target the stability and the variance of the current Bayesian parametric ROM.

The first acquisition function, $\psi:\mathcal{P}\to[0,1]$, is the probability of ROM instability, since an unstable surrogate model cannot be used for prediction or uncertainty estimation. Determining the stability of a probabilistic ROM analytically is challenging, but for a fixed parameter vector $\tilde{\vb*{\xi}}$, stability can be estimated empirically by drawing $n_d > 0$ ROM samples, integrating each sampled deterministic ROM in time at $\vb*{\xi} = \tilde{\vb*{\xi}}$, and counting the number of times $n_\alpha$ that the integration becomes numerically unstable. The probability of ROM instability, $\alpha = \psi(\tilde{\vb*{\xi}})\in [0,1]$, can then be estimated as
\begin{equation}
    \alpha = n_\alpha / n_d,
\end{equation}
which is a Monte Carlo estimator at $\vb*{\xi} = \tilde{\vb*{\xi}}$. This strategy is computationally tractable for moderate $n_d$ because each ROM sample is, by design, cheap to assemble and solve.

Our second acquisition function, $\varphi:\mathcal{P}\to\mathbb{R}_{\ge 0}$, targets global accuracy of the ROM by scoring parameter candidates based on the prediction uncertainties obtained from stable realizations of the Bayesian parametric ROM. We start by defining an ideal acquisition function based on the accuracy of the ROM as
\begin{align}
    \label{eq:acquisition-ideal}
    \varphi_\textrm{ideal}(\vb*{\xi})
    = \mathbb{E}\left[\frac{1}{T}\int_{0}^{T}\left\|
        \vb{q}(t;\vb*{\xi})
        - \breve{\vb{q}}_\textrm{rom}(t;\vb*{\xi})
        \right\|_{2}^{2}\,dt\right],
\end{align}
where $\vb{q}$ is the solution of the FOM and $\breve{\vb{q}}_\textrm{rom} = \vb{V}\hat{\vb{q}}_\textrm{rom}$ is the ROM solution mapped to the original state space. Maximizing $\varphi_\textrm{ideal}$ would find the point in the parameter domain where the ROM is least accurate in expectation. However, $\varphi_\textrm{ideal}(\vb*{\xi})$ cannot be computed without first computing the full-order solution $\vb{q}(t;\vb*{\xi})$. Instead, we derive an alternative based on the prediction variance of the Bayesian parametric ROM solution. For a vector random variable $\vb{x} = [~x_1~~\cdots~~x_{n_x}~]\trp\in\mathbb{R}^{n_x}$, recall the identity
\begin{align*}
    \mathbb{E}[\vb{x}\trp\vb{x}]
    = \mathbb{E}[\vb{x}]\trp\mathbb{E}[\vb{x}]
    + \operatorname{tr}\left(\operatorname{Cov}[\vb{x}]\right)
    = \left\|\mathbb{E}[\vb{x}]\right\|_2^2
    + \sum_{i=1}^{n_x} \operatorname{Var}[x_i].
\end{align*}
Suppressing for the moment the dependence of $\vb{q}$ and $\breve{\vb{q}}$ on $t$ and $\vb*{\xi}$,
\begin{align*}
    \varphi_\text{ideal}(\boldsymbol{\xi})
    &= \mathbb{E}\left[\frac{1}{T}\int_{0}^T\|\mathbf{q} - \breve{\mathbf{q}}\|_2^2\,dt\right]
\\
    &= \frac{1}{T}\int_{0}^T\mathbb{E}\left[(\mathbf{q} - \breve{\mathbf{q}})\trp(\mathbf{q} - \breve{\mathbf{q}})\right]\,dt
\\
    &= \frac{1}{T}\int_{0}^T \mathbf{q}\trp\mathbf{q} - 2\mathbf{q}\trp\mathbb{E}\left[\breve{\mathbf{q}}\right] + \mathbb{E}\left[\breve{\mathbf{q}}\trp\breve{\mathbf{q}}\right]\,dt
    \\
    &= \frac{1}{T}\int_{0}^T \mathbf{q}\trp\mathbf{q} - 2\mathbf{q}\trp\mathbb{E}\left[\breve{\mathbf{q}}\right] + \mathbb{E}\left[\breve{\mathbf{q}}\right]\trp\mathbb{E}\left[\breve{\mathbf{q}}\right] + \operatorname{tr}(\operatorname{Cov}[\breve{\mathbf{q}}])\,dt
\\
    &= \frac{1}{T}\int_{0}^T \|\mathbf{q} - \mathbb{E}\left[\breve{\mathbf{q}}\right]\|_2^2\,dt
    + \frac{1}{T}\int_{0}^T\sum_{i=1}^{n_x}\operatorname{Var}[\breve{q}_i]\,dt
    \\
    &= \varphi_\text{bias}(\vb*{\xi}) + \varphi_\text{var}(\vb*{\xi}),
\end{align*}
in which $\breve{q}_i(t;\vb*{\xi})$ the $i$-th entry of the ROM solution in the original state space, and where
\begin{align*}
    \varphi_\text{bias}(\vb*{\xi})
    &= \frac{1}{T}\int_{0}^T \|\mathbf{q}(t;\vb*{\xi}) - \mathbb{E}\left[\breve{\mathbf{q}}(t;\vb*{\xi})\right]\|_2^2\,dt,
    \\
    \varphi_\text{var}(\vb*{\xi})
    &= \frac{1}{T}\int_{0}^T\sum_{i=1}^{n_x}\operatorname{Var}[\breve{q}_i(t;\vb*{\xi})]\,dt.
\end{align*}
While $\varphi_\text{bias}(\vb*{\xi})$ depends on $\mathbf{q}(t;\vb*{\xi})$, $\varphi_\text{var}(\vb*{\xi})$ depends on ROM solutions only and can therefore be computed efficiently. Furthermore, letting $V_{i,k}$ denote the entries of the basis matrix $\vb{V}$, we have $\breve{q}_\textrm{i} = \sum_{k=1}^{r}V_{i,k}\hat{q}_{k}$, where $\hat{q}_{k}(t;\vb*{\xi})$ denotes the $k$-th entry of the vector-valued stochastic process $\hat{\vb{q}}_\textrm{rom}$. Then
\begin{align}
    \begin{aligned}
    \sum_{i=1}^{n_x}\Var[\breve{q}_{i}(t,\vb*{\xi})]
    &= \sum_{i=1}^{n_x}\sum_{k=1}^{r}\Var\big[
V_{i,k}\,\hat{q}_{k}(t;\vb*{\xi})\big]
    \\
    &= \sum_{i=1}^{n_x}\sum_{k=1}^{r}V_{i,k}^{2}\Var\big[\hat{q}_{k}(t;\vb*{\xi})\big]
= \sum_{k=1}^{r}\Var\big[\hat{q}_{k}(t;\vb*{\xi})\big],
\end{aligned}
    \label{eq:lowdimvariance}
\end{align}
since $\sum_{i=1}^{n_x}V_{i,k}^{2} = 1$. Hence, defining
\begin{align}
    \label{eq:acquisition-practical}
    \hat{\varphi}(\vb*{\xi})
    = \frac{1}{T}\int_{0}^{T}\sum_{k=1}^{r}\Var[\hat{q}_{k}(t,\vb*{\xi})]\dt,
\end{align}
we have $\hat{\varphi}(\vb*{\xi}) = \breve{\varphi}_\text{var}(\vb*{\xi})$. In other words, the total variance in the reduced space equals the total variance in the original state space. The acquisition function \cref{eq:acquisition-practical} is a computationally efficient measure of the total variance of the ROM solution at the given parameter vector, and this is the metric that we base our adaptive parameter selection strategy on.

\begin{remark}[Shifted approximation]
If the low-dimensional state approximation is given by $\vb{q} \approx \bar{\vb{q}} + \vb{V}\hat{\vb{q}}$ as in \Cref{remark:reduction}, the equality $\hat{\varphi}(\vb*{\xi}) = \varphi_\text{var}(\vb*{\xi})$ still holds because variance is translational. Specifically, we have $\breve{q}_{i} = \bar{q}_{i} + \sum_{k=1}^{r} V_{i,k}\,\hat{q}_{k}$, and $\Var\big[\bar{q}_{i} + V_{i,k}\,\hat{q}_{k}(t;\vb*{\xi})\big] = V_{i,k}^{2}\Var\big[\hat{q}_{k}(t;\vb*{\xi})\big]$, by which \cref{eq:lowdimvariance} holds.
\end{remark}

\Cref{alg:BayesOpInfAcquisition} provides a routine for scoring a parameter candidate in terms of stability and accuracy. Given draws from the operator matrix posterior, the corresponding deterministic ROMs are solved to produce state trajectories if possible or, otherwise, to indicate that the ROM is unstable. The probability of instability $\psi$ is estimated as $\alpha = n_\alpha / n_d$, and stable solutions are used to estimate the value of the variance-based acquisition function $\varphi$. In our experiments, we use a numerical approximant of \cref{eq:acquisition-practical},
\begin{align}
    \label{eq:acquisition-numerical}
    \varphi(\vb*{\xi})
    = \hat{\varphi}(\vb*{\xi})
    \approx \sum_{j=1}^{n_t}\sum_{k=1}^{r}\Var_{(\ell)}\big[\hat{q}_{k}^{(\ell)}(t_j;\vb*{\xi})\big],
\end{align}
in which $\hat{q}_{k}^{(\ell)}(t_j;\vb*{\xi})$ is the solution of the $\ell$-th deterministic ROM at time $t = t_j$ for some discretization of the time domain $t_1,t_2,\ldots,t_{n_t}$, and where $\Var_{(\ell)}$ indicates the sample variance over the sampling index $\ell$. The variance is only taken over stable draws (the set $S$ in \Cref{alg:BayesOpInfAcquisition}); consequently, this value cannot be computed if all deterministic ROM draws turn out to be unstable.

\subsection{Algorithm and implementation details}

\begin{algorithm}[t]
\begin{algorithmic}[1]
\Procedure{NextSample}{
    \newline\phantom{---}
    Parameter candidates $\tilde{\vb*{\xi}}_{1},\ldots,\tilde{\vb*{\xi}}_{\tilde{n}_{p}}\in \mathcal{P}$
\newline\phantom{---}
    Operator matrix realizations $\hat{\vb{O}}^{(1)},\ldots,\hat{\vb{O}}^{(n_d)}\in\mathbb{R}^{r\times d(r)}$
    \newline\phantom{---}
    Parametric ROM structure function $\vb{d} = \vb{d}(\hat{\vb{q}};\vb*{\xi})$
    \newline}
    \State $\mathcal{I} \gets \{1, 2, \ldots, \tilde{n}_{p}\}$
        \Comment{Indices of all candidates.}
    \For{$i = 1, \ldots, \tilde{n}_p$}
        \State $\alpha_i, \omega_i \gets \textproc{BayesOpInfAcquisition}(\tilde{\vb*{\xi}}_{i},\hat{\vb{O}}^{(1)},\ldots,\hat{\vb{O}}^{(n_d)},\vb{d})$
        \Comment{\Cref{alg:BayesOpInfAcquisition}.}
    \EndFor
    \State $\mathcal{U} \gets \{i\in\mathcal{I} \mid \alpha_{i} \ge \alpha_{j} ~\forall~ j \in \mathcal{I}\}$
        \Comment{Collect maximal unstable ROM indices.}
    \If{$\alpha_i < 100\%~\forall~i \in \mathcal{U}$ (some draws stable for every candidate)}
        \State $i^* \gets \operatorname{arg\,max}_{i\in\mathcal{U}} \omega_i$
            \Comment{Index maximizing acquisition.}
    \Else ~($\alpha_i = 100\%~\forall~i\in\mathcal{U}$, all draws unstable for at least one candidate)
        \State $i^* \gets$ sample randomly from $\mathcal{U}$
        \Comment{No stability.}
        \label{step:unstablerandomsample}
    \EndIf
\State \textbf{return} $\tilde{\vb*{\xi}}_{i^*}$
\EndProcedure
\end{algorithmic}
\caption{Adaptive sampling for (Bayesian) parametric OpInf ROMs}
\label{alg:BayesOpInfIteration}
\end{algorithm}

Let $\tilde{\vb*{\xi}}_{1},\ldots,\tilde{\vb*{\xi}}_{\tilde{n}_p}\in\mathcal{P}$ be a finite set of candidate parameter vectors at which we have the option to solve the FOM~\cref{eq:fom} to obtain training data for the pOpInf regression \cref{eq:popinf-both}. With an initial index $i_1 \in \mathcal{I} = \{1,\ldots,\tilde{n}_p\}$, our goal is to select a distinct index $i_2$ such that using training data corresponding to the parameters $(\vb*{\xi}_{1},\vb*{\xi}_{2}) = (\tilde{\vb*{\xi}}_{i_1},\tilde{\vb*{\xi}}_{i_2})$ results in the best possible pOpInf ROM among those that are trained with $\tilde{\vb*{\xi}}_{i_1}$ and one other parameter vector. We then seek an optimal third training parameter vector, and so on, until exhausting a computational budget for the FOM evaluations.

Our proposed procedure, which uses the stability and variance-based acquisition criteria from the previous section, is outlined in \Cref{alg:BayesOpInfIteration}. First, \Cref{alg:BayesOpInfAcquisition} is evaluated for each candidate parameter vector, resulting in pairs $(\alpha_i,\omega_i)$ for the probability of instability and variance-based acquisition values corresponding to each $\tilde{\vb*{\xi}}_{i}$. The indexing set $\mathcal{I}$ is downselected to those indices $\mathcal{U}$ corresponding to the parameter candidates that maximize the empirical probability of unstable behavior, that is, $\alpha_i \ge \alpha_j$ for all $i \in \mathcal{U}$ and $j \in \mathcal{I}$. From these remaining indices, the index $i^{*}$ is chosen which corresponds to the greatest acquisition value, meaning $\omega_{i^{*}} \ge \omega_{j}$ for all $j \in \mathcal{U}$. The procedure accounts for several scenarios:
\begin{enumerate}
\item If all ROM samples are stable for \emph{every} parameter candidate, then $\alpha_i = 0$ for each $i\in\mathcal{I}$ and therefore $\mathcal{U} = \mathcal{I}$. Hence, the selection of $i^{*}$ is determined exclusively using the acquisition values $\omega_{i}$, not via the stability indicators $\alpha_i$.

\item If there is a single parameter candidate index $i_{u}$ for which the instability indicator $\alpha_{i_u}$ is strictly greater than all other instability indicators, then $\mathcal{U} = \{i_u\}$ is a singleton set, and $i^{*} = i_u$ is always chosen no matter the corresponding variance-based acquisition value $\omega_{i_u}$. In other words, if there is a single ``most unstable'' parameter vector, the corresponding index is always chosen.

\item If there are several parameter candidate indices which maximize the stability indicator $\alpha_i$, then $\mathcal{U}$ is a subset of $\mathcal{I}$. This means that both acquisition functions are used---first stability, then variance.

\item If, for at least one parameter candidate, \emph{every} ROM sample is unstable, the variance-based acquisition value \cref{eq:acquisition-numerical} cannot be computed at that parameter (there are no stable solutions to take the variance of), and the winning index $i^{*}$ must be chosen randomly from the set $\mathcal{U}$ (\cref{step:unstablerandomsample}). From a practical point of view, this may indicate an issue with the problem setup, such as the quality of the low-dimensional state approximation or an inappropriate ROM structure.
\end{enumerate}

In the numerical experiments that follow, these cases tend to occur in reverse order:
with only few training parameter vectors, the ROM may be highly unstable at many parameter candidates; after sampling a few additional parameter vectors, the ROM tends to be unstable at only a few candidates; finally, given data for a sufficient number of training parameter vectors, the resulting ROM is stable throughout the domain and can then be improved strategically based on the variance-based acquisition function alone.

The entire active learning workflow is described in \Cref{alg:OuterLoop}. After generating an initial training set, a new parameter is selected via \Cref{alg:BayesOpInfIteration}, the basis and ROM operators are  recomputed, and the process repeats. The total number of FOM solves, which dominates the total computational cost, is the number of initial training parameters $n_p$ plus the number of times $n_b$ that the active learning loop is repeated.

\begin{algorithm}[t]
\begin{algorithmic}[1]
\Procedure{ActiveLearningROM}{
    \newline\phantom{---}
    Training parameter vectors $\vb*{\xi}_{1},\ldots,\vb*{\xi}_{n_p}\in\mathcal{P}$,
    \newline\phantom{---}
    Number of time steps $n_t$,
    \newline\phantom{---}
    Budget for additional full-order solves $n_b$,
    \newline\phantom{---}
    Parametric ROM structure function $\vb{d} = \vb{d}(\hat{\vb{q}};\vb*{\xi})$,
    \newline\phantom{---}
    Number of posterior draws $n_d > 0$,
    \newline\phantom{---}
    Parameter candidates $\tilde{\vb*{\xi}}_{1},\ldots,\tilde{\vb*{\xi}}_{\tilde{n}_{p}}\in \mathcal{P}$
    \newline}
    \LineComment{Generate initial training dataset.}
    \For{$i = 1, \ldots, n_p$}
        \State $\{\vb{q}_{ij}\}_{j=1}^{n_t} \gets$~Solve \cref{eq:generic-fom} with parameter value $\vb*{\xi}_i$ over $n_t$ time steps
    \EndFor
    \LineComment{Active learning loop.}
    \For{$k = 1, \ldots, n_b$}
\State $\bar{\vb{q}},\vb{V}\in\mathbb{R}^{N\times r} \gets$~Reference vector, basis matrix from $[~\vb{q}_{11}~\vb{q}_{12}~\cdots~\vb{q}_{n_pn_t}~]$
\State $\{\hat{\vb{q}}_{ij}\}_{i=1,j=1}^{n_p,n_t} \gets \{\vb{V}^{\mathsf{T}}(\vb{q}_{ij} - \bar{\vb{q}})\}_{i=1,j=1}^{n_p,n_t}$
            \Comment{Compress training data.}
        \State $\hat{\vb{O}}^{(1)},\ldots,\hat{\vb{O}}^{(n_d)}\gets\textproc{pBayesOpInf}(\{\vb*{\xi}_{i}\}_{i=1}^{n_p},\{\hat{\vb{q}}_{ij}\}_{i=1,j=1}^{n_p,n_t},\vb{d},n_d)$
            \Comment{Alg.~\ref{alg:pBayesOpInf}.}\State $\tilde{\vb*{\xi}}_{i^{*}} \gets \textproc{NextSample}(\{\tilde{\vb*{\xi}}_{i}\}_{i=1}^{\tilde{n}_p},\hat{\vb{O}}^{(1)},\ldots,\hat{\vb{O}}^{(n_d)},\vb{d})$
            \Comment{Alg.~\ref{alg:BayesOpInfIteration}.}\LineCommentt{Add selected parameter to training set.}
        \State $n_p \gets n_p + 1$
        \State $\vb*{\xi}_{n_p} \gets \tilde{\vb*{\xi}}_{i^{*}}$
        \State $\{\vb{q}_{n_{p}\,j}\}_{j=1}^{n_t} \gets$~Solve \cref{eq:generic-fom} with parameter value $\vb*{\xi}_{i^*}$ over $n_t$ time steps\LineCommentt{Remove selected parameter from candidate set.}
        \State $\{\tilde{\vb*{\xi}}_{i}\}_{i=1}^{\tilde{n}_{p}} \gets \{\ldots,\tilde{\vb*{\xi}}_{i^{*}-1},\tilde{\vb*{\xi}}_{i^{*}+1},\ldots\}$
        \State $\tilde{n}_p \gets \tilde{n}_{p} - 1$
    \EndFor
    \State \textbf{return} $\hat{\vb{O}}^{(1)},\ldots,\hat{\vb{O}}^{(n_d)}$
\EndProcedure
\end{algorithmic}
\caption{Active learning procedure for ROM learning.}
\label{alg:OuterLoop}
\end{algorithm}

\section{Numerical results}\label{sec:results}

In this section, we apply our active learning approach to learn probabilistic ROMs for two parametric partial differential equations (PDEs) with affine-parametric structure. Both examples are prototypical, yet they highlight the difficulty of constructing parametric ROMs that effectively capture the effects of the parameters on the solution, even with moderately benign parameter spaces, without an effective sampling strategy. The codes for these experiments are available at \href{https://github.com/McQLab/ActiveBayesOpInf}{\texttt{github.com/McQLab/ActiveBayesOpInf}}.

\subsection{Heat equation with nonlinear reaction}\label{sec:heat}

\begin{figure}[t]
    \centering
    \includegraphics[width=\textwidth]{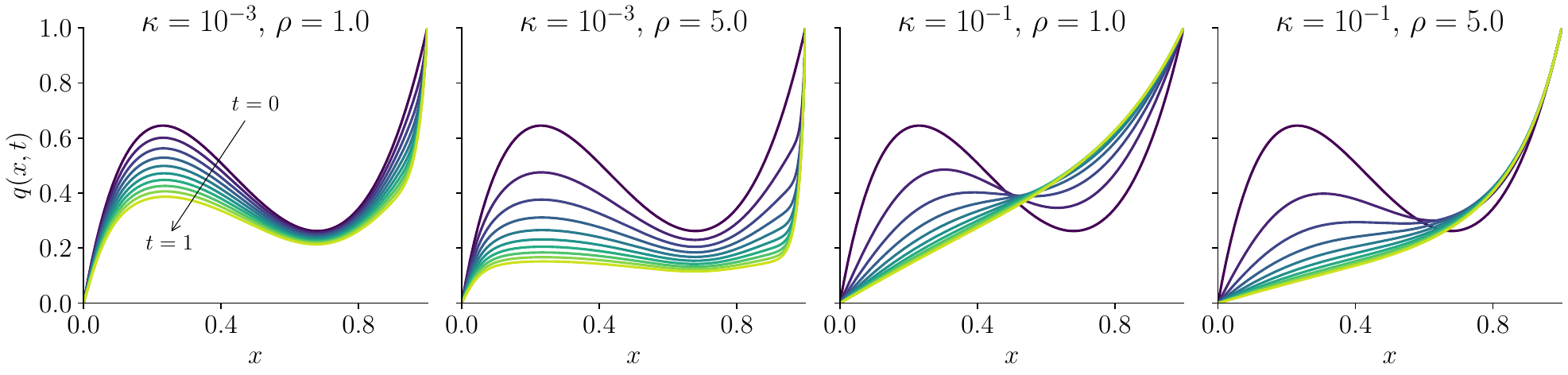}
    \caption{Solutions to the full-order diffusion-reaction model \cref{eq:heatfom} for values of the parameter vector $\vb*{\xi} = (\kappa,\rho)$ at the corners of the parameter domain $\mathcal{P} = [10^{-3}, 10^{-1}] \times [1, 5]$. The initial condition and time domain are the same for all experiments.}
    \label{fig:heat-snapshots}
\end{figure}

Consider the following diffusive PDE defined over a one-dimensional spatial domain,
\begin{align}
    \label{eq:heatpde}
    \begin{aligned}
    \frac{\partial q}{\partial t}
    &= \kappa \frac{\partial^{2} q}{\partial x^{2}} - \rho q^2,
    && x\in(0, L), t\in(0, t_f],
    \\
    q(0, t) &= 0,
    \quad
    q(L, t) = 1,
    && t\in(0, t_f],
    \\
    q(x, 0) &= q_{0}(x),
    && x\in [0, L],
    \end{aligned}
\end{align}
where $\kappa > 0$ and $\rho \ge 0$ are diffusion and reaction coefficients, respectively, $L > 0$ is the length of the spatial domain, $t_f > 0$ is the final time,
and $q_0:[0,L]\to\mathbb{R}$ is the initial condition. We consider a parameter-independent, non-polynomial initial condition that matches the constant Dirichlet boundary conditions,
\begin{align}
    q_0(x) = x(1 - x)(6 e^{-x} (1 - x)^2 - 10 e^x \sin(x / 6)) + x.
\end{align}
A uniform second-order central finite difference discretization of \cref{eq:heatpde} with $N$ degrees of freedom yields a semi-discrete model
\begin{align}
    \label{eq:heatfom}
    \frac{\text{d}\vb{q}}{\text{d}t}
    = \kappa\vb{c}^{(1)}
    + \kappa\vb{A}^{\!(1)}\vb{q}
    + \rho\vb{H}^{(1)}[\vb{q}\otimes\vb{q}],
\end{align}
where $\vb{q}(t;\vb*{\xi}),\vb{c}^{(1)}\in\mathbb{R}^{N}$, $\vb{A}^{\!(1)}\in\mathbb{R}^{N\times N},$ and $\vb{H}^{(1)}\in\mathbb{R}^{N\times N^2}.$
This is the generic FOM \cref{eq:fom} with parameter vector $\vb*{\xi} = (\kappa, \rho)$, coefficient functions $\theta_c^{(1)}\!(\vb*{\xi}) = \theta_a^{(1)}\!(\vb*{\xi}) = \kappa$ and $\theta_h^{(1)}\!(\vb*{\xi}) = \rho$, with $n_c = n_a = n_h = 1$. In our experiments, we use $L = 1$ as the length of the spatial domain, set $N = 500$, and consider the time domain $t \in [0, 1]$ with $n_t = 101$ discrete temporal points.
\Cref{fig:heat-snapshots} shows FOM solutions for several parameter vectors.

Our goal in this experiment is to construct a ROM that accurately mirrors the FOM over the parameter domain $(\kappa,\rho) \in \mathcal{P} = [10^{-3}, 10^{-1}] \times [1, 5]$. To exactly preserve the boundary conditions, we consider affine approximations of the discretized state, $\vb{q} \approx \bar{\vb{q}} + \vb{V}\hat{\vb{q}}$, where $\bar{\vb{q}}$ is a reference vector such that $\vb{q}(t) - \bar{\vb{q}}$ corresponds to homogeneous Dirichlet conditions. The corresponding ROM has the form
\begin{align}
    \begin{aligned}
    \frac{\text{d}\hat{\vb{q}}}{\text{d}t}
    = \kappa\hat{\vb{c}}^{(1)}
    + \rho\hat{\vb{c}}^{(2)}
    + \left(\kappa\hat{\vb{A}}^{\!(1)} + \rho\hat{\vb{A}}^{\!(2)}\right)\hat{\vb{q}}
    + \rho\hat{\vb{H}}^{(1)}[\hat{\vb{q}}\otimes\hat{\vb{q}}],
    \end{aligned}
\end{align}
where
$\hat{\vb{c}}^{(1)},\hat{\vb{c}}^{(2)}\in\mathbb{R}^{r}$, $\hat{\vb{A}}^{\!(1)},\hat{\vb{A}}^{\!(2)}\in\mathbb{R}^{r\times r}$, and $\hat{\vb{H}}^{(1)}\in\mathbb{R}^{r\times r^2}$. This is the generic ROM \cref{eq:rom} with coefficient functions inherited from \cref{eq:heatfom}, plus additional coefficient functions $\theta_{c}^{(2)}\!(\vb*{\xi}) = \theta_{a}^{(2)}\!(\vb*{\xi}) = \rho$ arising due to the nonzero reference vector $\bar{\vb{q}}$. To write this as a pOpInf ROM $\dxdt{\hat{\vb{q}}} = \hat{\vb{O}}\vb{d}(\hat{\vb{q}},\vb*{\xi})$ akin to \cref{eq:opinf-rom}, we define
\begin{align}
    \label{eq:heat-Ohat}
    \hat{\vb{O}}
    &= [~\hat{\vb{c}}^{(1)}~~\hat{\vb{c}}^{(2)}~~\hat{\vb{A}}^{\!(1)}~~\hat{\vb{A}}^{\!(2)}~~\hat{\vb{H}}^{(1)}~]
    \in\mathbb{R}^{r\times d(r)},
    &
    \vb{d}(\hat{\vb{q}},\vb*{\xi})
    &= \left[\begin{array}{c}
        \kappa \\
        \rho \\
        \kappa\hat{\vb{q}} \\
        \rho\hat{\vb{q}} \\
        \rho(\hat{\vb{q}}\otimes\hat{\vb{q}})
    \end{array}\right]
    \in\mathbb{R}^{d(r)},
\end{align}
with data dimension $d(r) = 2 + 2r + r^2$.

\begin{remark}[Compressed Kronecker product]
The quadratic mapping $\hat{\vb{q}} \mapsto \hat{\vb{H}}^{(1)}[\hat{\vb{q}} \otimes \hat{\vb{q}}]$ with $\hat{\vb{H}}\in\mathbb{R}^{r \times r^2}$ can be represented with a compressed version of the Kronecker product and a $r \times r(r+1)/2$ matrix to avoid redundant computations, see, e.g., \cite[Appendix~B]{mcquarrie2023thesis} for details. Hence, in practice, the data dimension for~\cref{eq:heat-Ohat} is reduced to $d(r) = 2 + 2r + r(r+1)/2$.
\end{remark}

To assess our adaptive sampling strategy, we discretize the two-dimensional parameter domain $\mathcal{P}$ with a $20\times 20$ grid $\mathcal{G}\subset\mathcal{P}$ of parameter candidates, logarithmically spaced in $\kappa$ and uniformly spaced in $\rho$. We run $50$ trials where, in each trial, a single parameter sample is randomly selected from the parameter grid $\mathcal{G}$ to generate an initial training data set via the FOM. From that data, we learn a Bayesian parametric ROM with the procedure from \Cref{sec:opinf}, then select a new training parameter vector using the sampling strategy of \Cref{sec:implementation}. The FOM is then solved at the new parameter sample to generate additional training data, a new ROM is constructed, and another sample is selected. This procedure continues until, in each trial, $n_p = 10$ training parameters have been selected. A new POD basis matrix $\vb{V}$ is generated at each iteration by calculating SVD of the (shifted) snapshot matrix with mean-zero columns $\vb{q}_{j} - \bar{\vb{q}}$, where the $\vb{q}_{j}$ are the training snapshots and $\bar{\vb{q}}$ is the average training snapshot. Letting $\sigma_{1} \ge \sigma_{2} \ge \cdots \ge \sigma_{K}$ be the resulting singular values, the reduced dimension $r$ is selected as the smallest integer such that the residual singular value energy,
\begin{align*}
1 - \left(\sum_{i=1}^{r}\sigma_{i}^{2}
    \middle/
    \sum_{k=1}^{K}\sigma_{k}^{2}\right)
    = \left.\sum_{i=r+1}^{K}\sigma_{i}^{2}
    \middle/
    \sum_{k=1}^{K}\sigma_{k}^{2}\right.,
\end{align*}
is less than $10^{-6}$. The POD basis matrix $\vb{V}$ then consists of the first $r$ left singular vectors.

\begin{figure}[t]
    \centering
    \includegraphics[width=\textwidth]{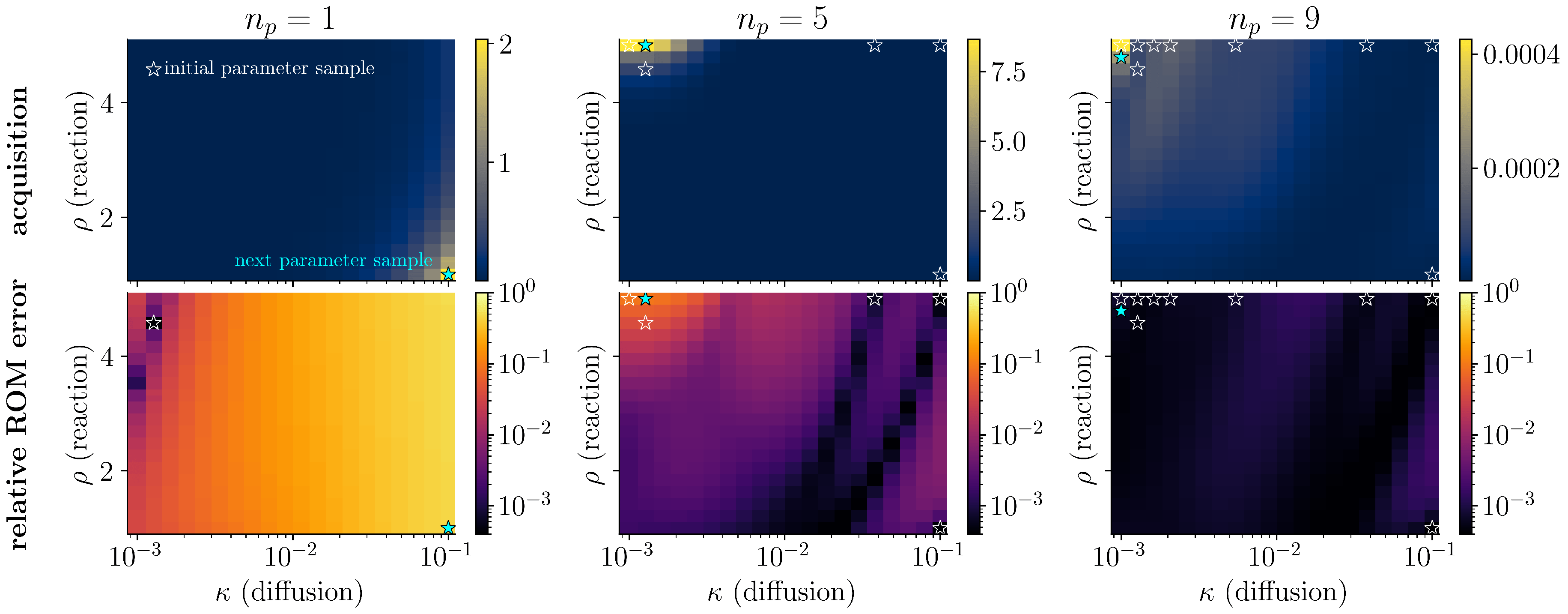}
    \caption{Acquisition values $\varphi(\vb*{\xi})$ (top) and relative $L^2$-norm ROM errors (bottom) after training on $n_p$ adaptively chosen parameter samples for the problem \cref{eq:heatpde}.}
    \label{fig:heat-iteration}
\end{figure}

\begin{remark}[Unstructured candidate sets]
\label{remark:grids}
The set of parameter candidates $\mathcal{G}$ is chosen in this example to be a regular grid in order to facilitate visualizing the results and establish a fair comparison with realistic random sampling (described shortly). However, the set $\mathcal{G}$ need not represent a grid or any other kind of geometric structure; all that \Cref{alg:OuterLoop} requires is a generic set of candidates to evaluate and select from.
\end{remark}

At each iteration of the procedure,
we evaluate the learned ROM on each of the $400$ parameter vectors in the grid $\mathcal{G}$ and compute the $L^2$ error relative to the FOM, which is related to the ideal acquisition function $\varphi_\text{ideal}$ in \cref{eq:acquisition-ideal}. Specifically, the relative ROM error is defined by
\begin{align}
    \label{eq:relative-error}
    \text{\bf relative ROM error}
= \frac{
        \int_{0}^{T}\|\vb{q}(t; \vb*{\xi}) - \mathbb{E}[\breve{\vb{q}}_\text{rom}(t;\vb*{\xi})]\|_{2}^{2}\,dt
    }{
        \int_{0}^{T}\|\vb{q}(t; \vb*{\xi})\|_{2}^{2}\,dt
    },
\end{align}
which is a function of the parameter vector $\vb*{\xi}$. The time integrals are estimated with the trapezoidal rule over the discrete time domain and the expectation is estimated with a sample mean over $n_d = 50$ draws. Recall that this error is unavailable without evaluating the FOM; our goal is for this error to decrease globally---without our having access to it---as parameter samples are strategically added to the training set.

\Cref{fig:heat-iteration} displays the relative ROM error after training on data corresponding to $n_p = 1$, $n_p = 5$, and $n_p = 9$ parameter samples for a single trial. The ROM error tends to be smallest near training parameter samples and is often largest at the boundaries of the parameter domain.
\Cref{fig:heat-iteration} also shows the variance-based acquisition values $\varphi(\vb*{\xi})$, which are used to adaptively determine the next parameter sample. Although the acquisition function is based on variance, not actual error, the parameters that maximize the acquisition function tend to also maximize or nearly maximize the ROM error. Hence, the adaptive sampling procedure successfully selects parameter vectors where the current ROM performs poorly. As $n_p$ increases, the ROM error decreases globally across $\mathcal{P}$. The variance-based acquisition values do not always decrease monotonically as new parameter samples are added, but these values are typically much smaller for larger $n_p$ than for smaller $n_p$. This indicates that as the ROM becomes more accurate, it also reports more confidence in its predictions.

\begin{figure}[t]
    \centering
    \includegraphics[width=\textwidth]{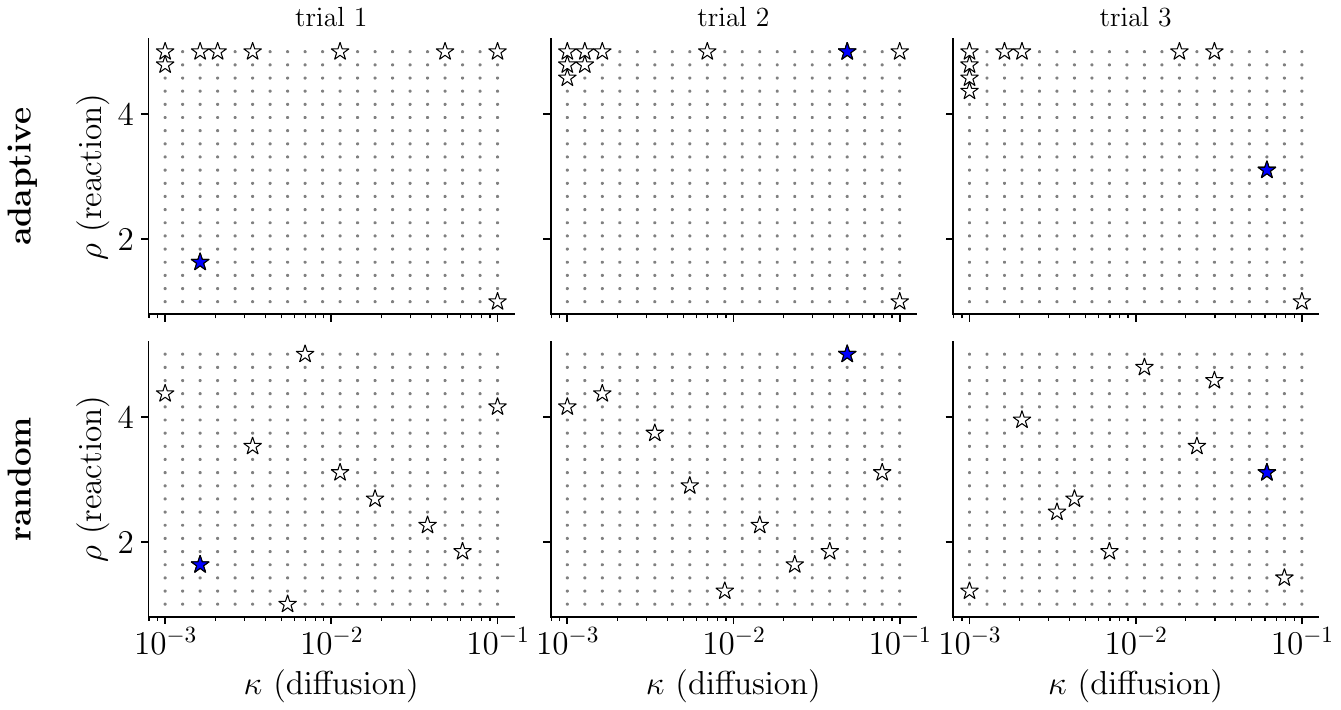}
    \caption{Parameter samples selected from a $20\times 20$ grid $\mathcal{G}$ over the parameter domain $\mathcal{P}$ using adaptive sampling (top) and Latin hypercube sampling (bottom) for the problem \cref{eq:heatpde}. Results are shown for three trials, each with a different random initial training parameter vector (blue).}
    \label{fig:heat-samples}
\end{figure}

We also compare our adaptive sampling procedure to random sampling. For each of the $50$ experimental trials (each with different initial training samples), we select $n_p = 10$ training parameters from the $20\times 20$ grid $\mathcal{G}\subset\mathcal{P}$ via discrete Latin hypercube sampling while ensuring that the initial samples for each trial are the same as were used in the corresponding adaptive sampling experiments.
\Cref{fig:heat-samples} shows the samples from three different trials for both the adaptive sampling strategy and random sampling. The Latin hypercube sampling covers the space $\mathcal{P}$ better, while the adaptively chosen samples tend toward the corners and boundaries of $\mathcal{P}$. The exact training parameters chosen by the adaptive sampling vary from trial to trial, depending on the initial parameter sample, but many adaptive samples are always chosen from the region of $\mathcal{P}$ with the largest reaction coefficient $\rho$ and, especially, with large $\rho$ but small diffusion coefficient $\kappa$. Note that this corner of $\mathcal{P}$ is where the model~\cref{eq:heatpde} is the most nonlinear, since $\rho$ is attached to the reaction term and $\kappa$ scales the linear diffusion operator. On the other hand, the adaptive sampling scheme almost always also selects one sample from the opposite corner of $\mathcal{P}$, where $\kappa$ is largest and $\rho$ is smallest---the area of $\mathcal{P}$ where \cref{eq:heatpde} is the most linear. Clearly, the adaptive sampling scheme avoids replicating uniform or space-covering sampling, instead focusing on particular regions of the parameter domain.

\begin{figure}[t]
    \centering
    \includegraphics[width=\textwidth]{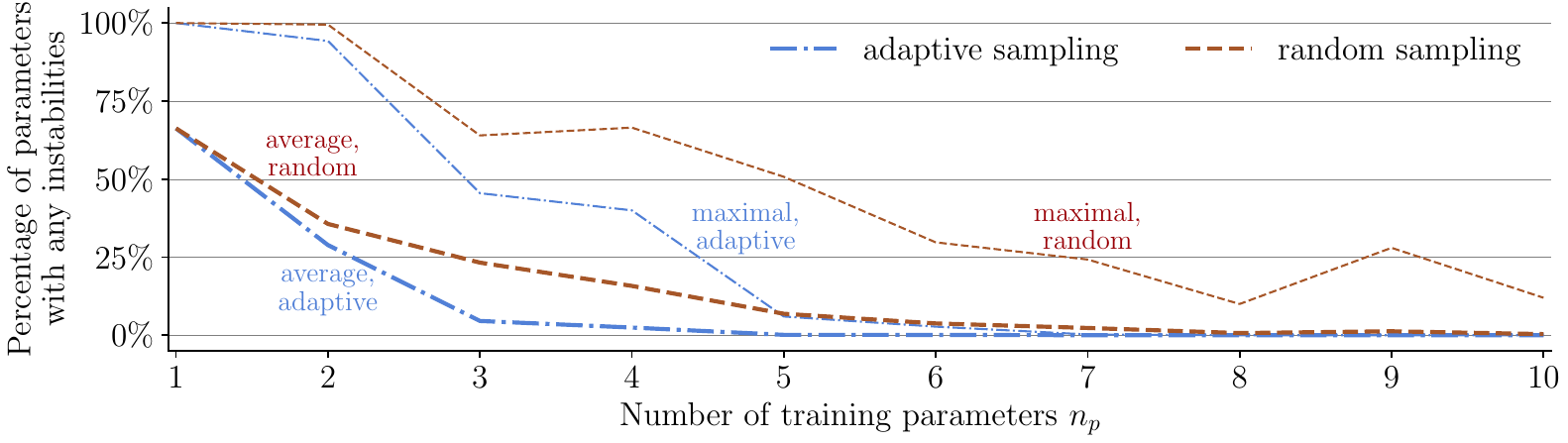}
    \caption{Percentages of the $400$ training parameter candidates where any instabilities are observed from $50$ posterior draws of the probabilistic ROM, as a function of the number of training parameters $n_p$, for the problem \cref{eq:heatpde}. The results are averaged and maximized over $50$ trials with different random initial parameter samples.}
    \label{fig:heat-instability}
\end{figure}

\begin{figure}[t]
    \centering
    \includegraphics[width=\textwidth]{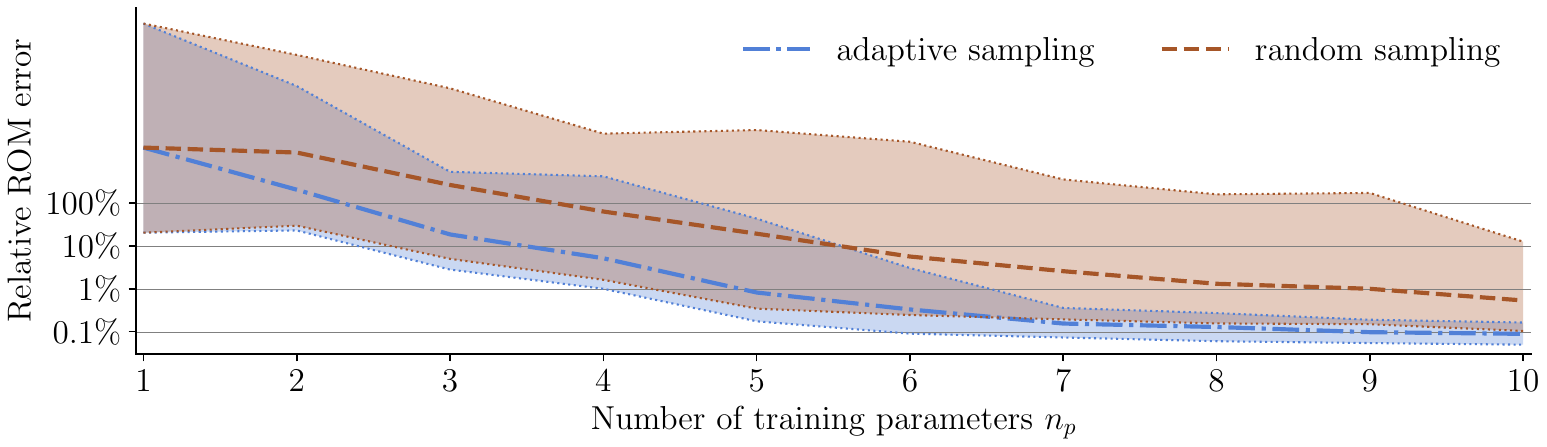}
    \caption{Total relative $L^2$ ROM errors $\mathcal{E}_\textrm{total}$ as a function of the number of training parameter samples, aggregated across 50 trials with different initial samples, for the problem \cref{eq:heatpde}. The shaded regions show the interquantile range of $5\%$--$95\%$ of trials.
    }
    \label{fig:heat-romerror}
\end{figure}

\begin{figure}[t]
    \centering
    \includegraphics[width=\textwidth]{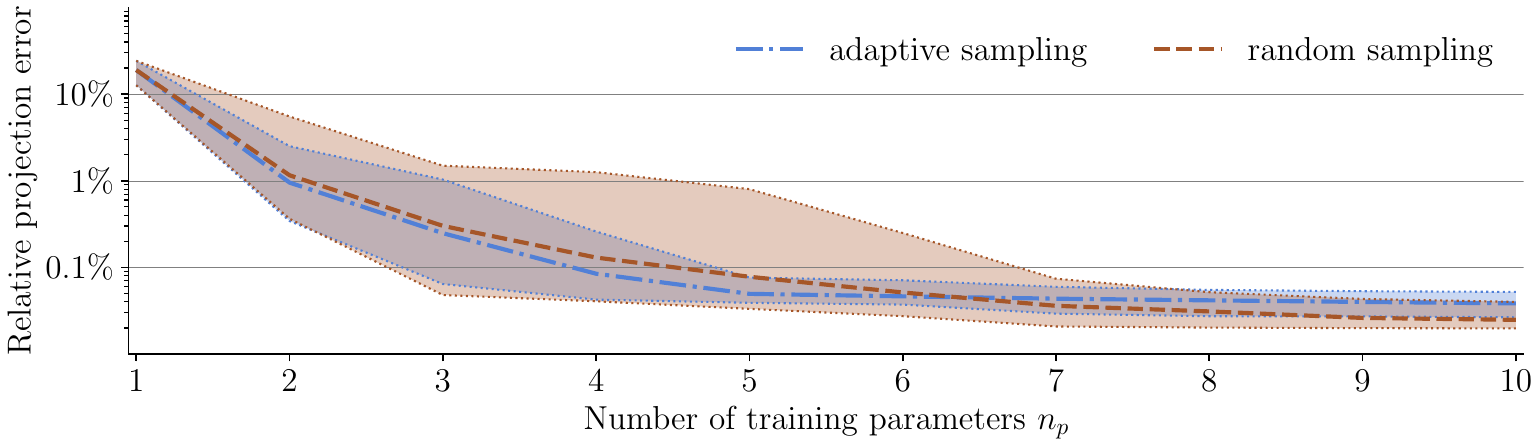}
    \caption{Total relative $L^2$ projection errors $\mathcal{E}_\textrm{proj}$ as a function of the number of training parameter samples, aggregated across 50 trials with different initial samples, for the problem \cref{eq:heatpde}. The shaded regions show the interquantile range of $5\%$--$95\%$ of trials.}
    \label{fig:heat-projectionerror}
\end{figure}

Next, we discuss how the sampling strategies compare in terms of ROM stability. In each trial and at each iteration, we record the probability of ROM instability $\alpha_{i} = \psi(\vb*{\xi}_{i}) \in [0, 1]$ for each parameter candidate $\vb*{\xi}_{i}$ in the grid $\mathcal{G}$. We then calculate the percentage of parameters for which $\alpha_{i} > 0$, that is, the frequency with which any of the posterior ROM draws were unstable at a given parameter vector:
\begin{align*}
    \beta = \frac{1}{400}\sum_{i=1}^{400}\beta_{i},
    \qquad\beta_i = \begin{cases}
        1 & \text{if}~\alpha_i > 0, \\
        0 & \text{if}~\alpha_i = 0.
    \end{cases}
\end{align*}
\Cref{fig:heat-instability} shows these $\beta$ values as a function of the number of training parameters $n_p$, averaged over the $50$ trials for each sampling method. We also display the maximum over the trials (pointwise for each $n_p$). With only one training parameter sample, the resulting parametric ROM is unstable (for at least one posterior realization) over about $65\%$ of the parameter grid on average, while at least one initial parameter sample resulted in a ROM that was unstable over all of $\mathcal{G}$. There are two key insights from \Cref{fig:heat-instability}: (i) the average and maximal instability percentages decrease faster with adaptive sampling than with random sampling, and (ii) adaptive sampling achieves total stability for $n_p \ge 7$, while at least one of the trials for random sampling results in some instabilities for all $n_p \le 10$. With $n_p = 6$, random sampling often results in ROMs that are stable over the entire parameter domain, but the same average rate of stability can be achieved with half as many samples through adaptive sampling.

We conclude this example by analyzing the accuracy of the expected value of ROM predictions.
\Cref{fig:heat-iteration} plots the relative error \cref{eq:relative-error} over the parameter domain for a few select cases; to measure the ROM error over the entire parameter space, we consider a total $L^2$ error,
\begin{align}
    \label{eq:total-error}
    \mathcal{E}_\textrm{total}
    = \left(\frac{
        \sum_{\vb*{\xi}_{i}\in\mathcal{G}}\left(
            \int_{0}^{T}\|\vb{q}(t; \vb*{\xi}_{i}) - \mathbb{E}[\breve{\vb{q}}_\text{rom}(t;\vb*{\xi}_{i})]\|_{2}^{2}\,dt
        \right)^{2}
    }{
        \sum_{\vb*{\xi}_{i}\in\mathcal{G}}\left(
            \int_{0}^{T}\|\vb{q}(t; \vb*{\xi}_{i})\|_{2}^{2}\,dt
        \right)^{2}
    }\right)^{1/2}.
\end{align}
This value is calculated for each of the $50$ trials, and the results are aggregated in \Cref{fig:heat-romerror}, which shows the $5\%$ and $95\%$ quantiles and the geometric mean of the $\mathcal{E}_\textrm{total}$ values. After only $n_p=2$ training samples, the adaptive sampling ROMs perform at least one order of magnitude better on average than the random sampling ROMs. Furthermore, the range of errors contracts quickly in the adaptive sampling case, so that with $n_p = 7$ training samples the ROM error is always below $1\%$ (and sometimes below $0.1\%$). By contrast, in the random sampling case, the ROM error spans a wide range of at least three orders of magnitude for all $n_p \le 9$. Even with $n_p = 10$ samples, in at least one instance the random sampling ROM still only achieves $10\%$ total relative error. The difference in the results shown in \Cref{fig:heat-romerror} is primarily due to the manner in which the adaptive sampling method targets ROM performance, \emph{not} because of the quality of the underlying basis. To demonstrate this, \Cref{fig:heat-projectionerror} shows the total projection error,
\begin{align}
    \label{eq:total-projectionerror}
    \mathcal{E}_\textrm{proj}
    = \left(\frac{
        \sum_{\vb*{\xi}_{i}\in\mathcal{G}}\left(
            \int_{0}^{T}\|\vb{q}(t; \vb*{\xi}_{i}) - \vb{V}\vb{V}\trp\left(\vb{q}(t;\vb*{\xi}_{i}) - \bar{\vb{q}}\right) + \bar{\vb{q}}\|_{2}^{2}\,dt
        \right)^{2}
    }{
        \sum_{\vb*{\xi}_{i}\in\mathcal{G}}\left(
            \int_{0}^{T}\|\vb{q}(t; \vb*{\xi}_{i})\|_{2}^{2}\,dt
        \right)^{2}
    }\right)^{1/2},
\end{align}
which depends only on the basis matrix $\vb{V}$, not the quality of the ensuing ROM. Recall that the basis size is determined at each iteration by the residual singular value energy. This means that the basis may grow when derived from a more diverse set of training data: in these experiments, the initial basis with $n_p = 1$ training parameter has $r = 4$ or $5$, which increases to about $r = 8$ for $n_p = 2$, and settles at $r = 10$ or $r = 11$ after additional samples. This is the main reason that the projection error first decreases, then plateaus, as $n_p$ increases. The average rate of decrease is highly similar in adaptive and random sampling, and the variance in the projection errors is smaller for the adaptive sampling, which tends to focus on the same regions of $\mathcal{P}$ in each trial. This shows that the adaptive sampling method does not merely target a better basis matrix $\vb{V}$; rather, it focuses specifically on improving the performance of the resulting ROM.

\subsection{Viscous Burgers' equation}\label{sec:burgers}

\begin{figure}[t]
    \centering
    \includegraphics[width=\linewidth]{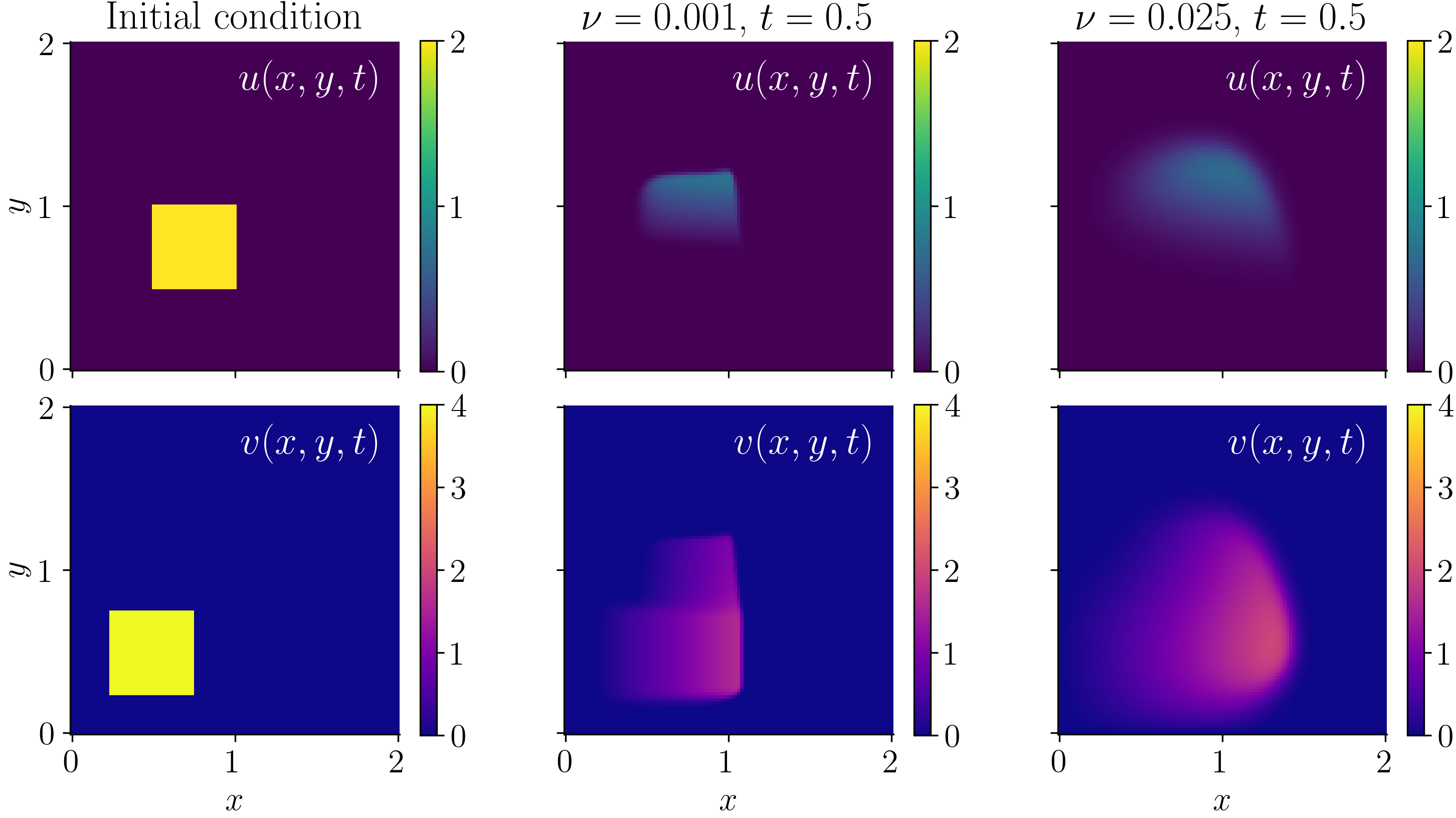}
    \caption{Solutions to the full-order Burgers' equation \cref{eq:burgers2D} at the initial time (left) and for two different values of $\nu$ at time $t = 0.5$ (middle and right).
    With smaller $\nu$, the solutions are more shock-like (middle); when $\nu$ is larger, the solutions are more diffusive (right).}
    \label{fig:burgers-snapshots}
\end{figure}

Consider next the two-dimensional viscous Burgers’ equations over the spatial domain $\Omega = [0, 2] \times [0, 2]$,
\begin{subequations}\label{eq:burgers2D}
\begin{align}
    \label{eq:burgers2D-pde}
    \begin{aligned}
    \frac{\partial u}{\partial t} + u \frac{\partial u}{\partial x} + v \frac{\partial u}{\partial y}
    &= \nu \left( \frac{\partial^2 u}{\partial x^2} + \frac{\partial^2 u}{\partial y^2} \right),
    \\
    \frac{\partial v}{\partial t} + u \frac{\partial v}{\partial x} + v \frac{\partial y}{\partial y}
    &= \nu \left( \frac{\partial^2 v}{\partial x^2} + \frac{\partial^2 v}{\partial y^2} \right),
    \end{aligned}
\end{align}
where $u(x, y, t)$ and $v(x, y, t)$ are the velocity in the $x$ and $y$ directions, respectively, and $\nu \in \mathcal{P} = \{0.001,0.025\}$ is the kinematic viscosity, which is the parameter of interest. We consider homogeneous Dirichlet boundary conditions,
\begin{align}
    \label{eq:burgers-bcs}
    u(x,y,t) &= 0,
    \qquad
    v(x,y,t) = 0,
    \qquad
    \forall(x,y)\in\partial\Omega, \, t\ge 0,
\end{align}
and define the initial velocity fields as partially overlapping square pulses:
\begin{align}
    \label{eq:burgers2D-initialconditions}
    \begin{aligned}
    u(x, y, 0) &=
    \begin{cases}
    2, & 0.5 \le x \le 1.0,\ 0.5 \le y \le 1.0, \\
    0, & \text{elsewhere},
    \end{cases}
    \\
    v(x, y, 0) &=
    \begin{cases}
    4, & 0.25 \le x \le 0.75,\ 0.25 \le y \le 0.75, \\
    0, & \text{elsewhere}.
    \end{cases}
    \end{aligned}
\end{align}
\end{subequations}
This setup produces a localized velocity disturbance that spreads and diffuses over time, with stronger diffusion when $\nu$ is larger and more shock-like behavior for small $\nu$.

Discretizing \cref{eq:burgers2D-pde} with finite differences over $n_x$ spatial degrees of freedom yields a semi-discrete model
\begin{align}
    \label{eq:burgersfom}
    \frac{\text{d}\vb{q}}{\text{d}t}
    = \nu\vb{A}^{\!(1)}\vb{q}
    + \vb{H}^{(1)}[\vb{q}\otimes\vb{q}],
    \qquad
    \vb{q} = \left[\begin{array}{c}
        \vb{u} \\ \vb{v}
    \end{array}\right],
\end{align}
in which $\vb{u}(t;\nu),\vb{v}(t;\nu)\in\mathbb{R}^{n_x}$ are the spatial discretizations of $u$ and $v$, respectively, and where $\vb{q}(t;\nu)\in\mathbb{R}^{N}$, $\vb{A}^{(1)} \in \mathbb{R}^{N \times N}$ and $\vb{H}^{(1)} \in \mathbb{R}^{N\times N^2}$ with $N = 2n_x$. This is the FOM \cref{eq:fom} with parameter ``vector'' $\vb*{\xi} = \nu$, coefficient functions $\theta_{a}^{(1)}\!(\vb*{\xi}) = \nu$ and $\theta_{h}^{(1)}\!(\vb*{\xi}) = 1$, with $n_c = 0$ (no constant term) and $n_a = n_h = 1$. Our experiments use $n_x = 101^2 = 10{,}201$ ($101$ uniformly spaced points in each spatial direction) so that the FOM has $N = 2n_x = 20{,}402$ total degrees of freedom, along with the time domain $t \in [0,1]$ with $n_t = 100$ temporal measurements.
\Cref{fig:burgers-snapshots} shows the initial conditions \cref{eq:burgers2D-initialconditions} and solutions to the FOM \cref{eq:burgersfom} at $t = 0.5$ for the smallest and largest values of $\nu$. Although this problem has only a one-dimensional parameter space, it is more difficult to tackle with standard model reduction methods than the previous problem due to the shock-like behavior that occurs with small values of $\nu$.

Because the boundary conditions \cref{eq:burgers-bcs} are homogeneous, we consider a linear low-dimensional approximation of the discrete state, $\vb{q} \approx \vb{V}\hat{\vb{q}}$ (i.e., the reference vector is $\bar{\vb{q}} = \vb{0}$). This leads to a ROM which inherits the parametric and polynomial structure of the FOM exactly,
\begin{align}
    \label{eq:burgersrom}
    \frac{\text{d}\hat{\vb{q}}}{\text{d}t}
    = \nu\hat{\vb{A}}^{\!(1)}\hat{\vb{q}}
    + \hat{\vb{H}}^{(1)}[\hat{\vb{q}}\otimes\hat{\vb{q}}],
\end{align}
where $\hat{\vb{A}}^{(1)}\in\mathbb{R}^{r\times r}$, $\hat{\vb{H}}^{(1)}\in\mathbb{R}^{r\times r^2}$. This ROM can be written in the form of \cref{eq:opinf-rom} by defining
\begin{align}
    \hat{\vb{O}}
    = [~\hat{\vb{A}}^{(1)}~~\hat{\vb{H}}^{(1)}~]
    \in\mathbb{R}^{r\times d(r)},
    \qquad
    \vb{d}(\hat{\vb{q}},\vb*{\xi})
    = \left[\begin{array}{c}
        \nu\hat{\vb{q}} \\ \hat{\vb{q}}\otimes\hat{\vb{q}}
    \end{array}\right]
    \in\mathbb{R}^{d(r)},
\end{align}
where the data dimension is $d(r) = r + r^2$.

For our experiments, we discretize the one-dimensional parameter domain $\mathcal{P}$ with a uniformly spaced grid $\mathcal{G}\subset\mathcal{P}$ of $50$ logarithmically spaced parameter candidates. As in the previous experiment, we run $50$ trials, each with a different initial condition, in which we iteratively learn a Bayesian parametric ROM, adaptively select a new training parameter value, and repeat until obtaining $n_p = 10$ samples. The POD basis matrix $\vb{V}$ is constructed  with the $r$ principal left singular vectors of the snapshot data where $r$ is the smallest integer such that the cumulative energy
\begin{align*}
    \left.\sum_{i=1}^{r}\sigma_{i}^{2}
    \middle/
    \sum_{k=1}^{K}\sigma_{k}^{2}\right.
\end{align*}
exceeds $99.5\%$, where $\sigma_{1}\ge\sigma_{2}\ge\cdots\ge\sigma_{K}$ are the singular values of the snapshot matrix, as before. The following $L^1$-$L^2$ norm is appropriate for the shock-like phenomena in this problem:
\begin{align}
    \left\|q\right\|_{L^1(\Omega)^2} =
    \left\|~\left\|\begin{array}{c} u \\ v \end{array}\right\|_2\right\|_{L^1(\Omega)}
= \int_\Omega \sqrt{u(x)^2 + v(x)^2} dx,
\end{align}
which we use to define the relative ROM error, total relative error, and relative projection error analogous to \cref{eq:relative-error}, \cref{eq:total-error}, and \cref{eq:total-projectionerror}, respectively.

\begin{figure}[t]
    \centering
    \includegraphics[width=\linewidth]{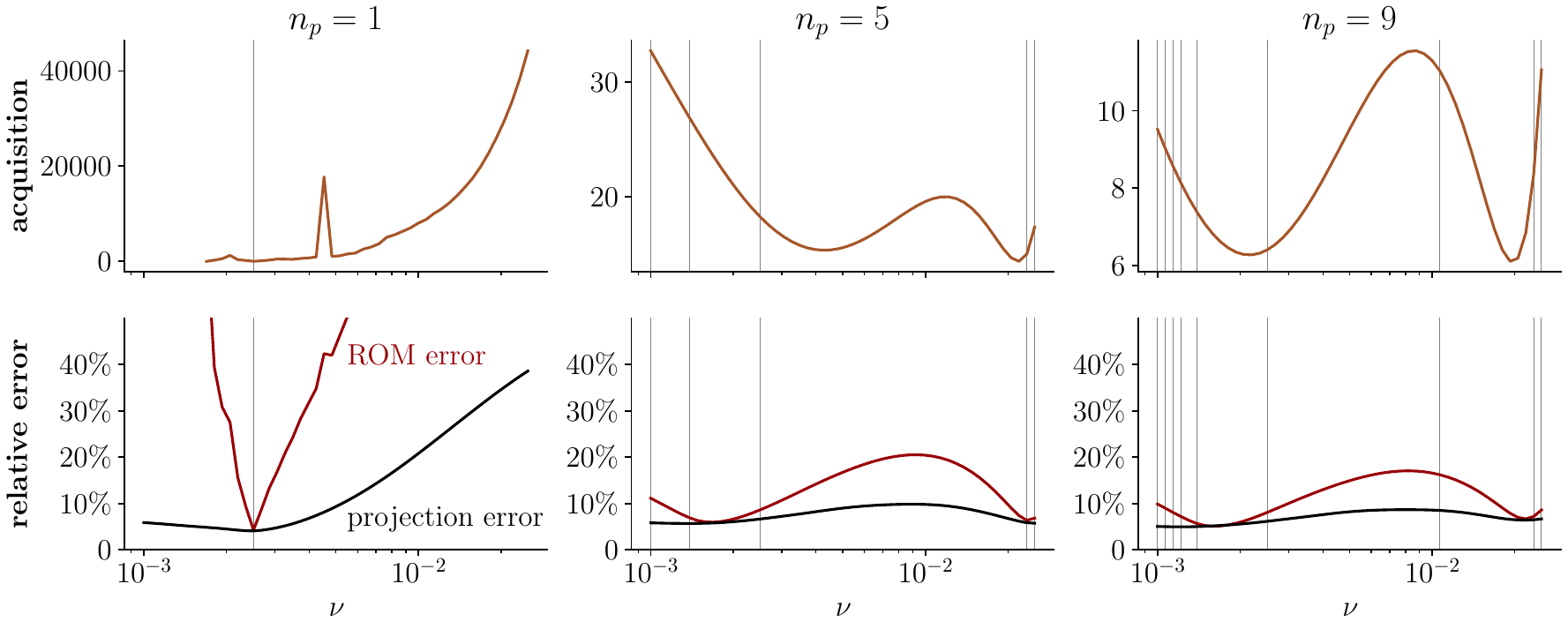}
    \caption{Acquisition values $\varphi(\nu)$ (top) and relative ROM and projection errors (bottom) after training on $n_p$ adaptively chosen parameter samples in the problem \cref{eq:burgers2D}. The vertical lines show the locations of the parameter samples.}
    \label{fig:burgers-iteration}
\end{figure}

\Cref{fig:burgers-iteration} shows the relative ROM and projection errors and the variance-based acquisition values in a single trial when there are $n_p=1,5,$ and $9$ training parameter values. In this representative trial, when $n_p = 1$ the ROM error is very near the projection error at the solitary training parameter value, but the ROM (and the basis) generalizes poorly to other parameter values. After additional parameter samples, however, both the projection error and the ROM error drop to a reasonable range throughout $\mathcal{P}$, with the ROM nearly achieving the projection error at some (but not all) parameter values. The adaptive sampling scheme tends to select parameters near the boundaries of the parameter domain, though there are exceptions. As the number of training parameter samples $n_p$ increases, the acquisition values decrease and become better indicators for the trends in the ROM error. Thus, with more data the ROM becomes more confident in general but is also equipped with an informative uncertainty indicator.

\begin{figure}[t]
    \centering
    \includegraphics[width=\textwidth]{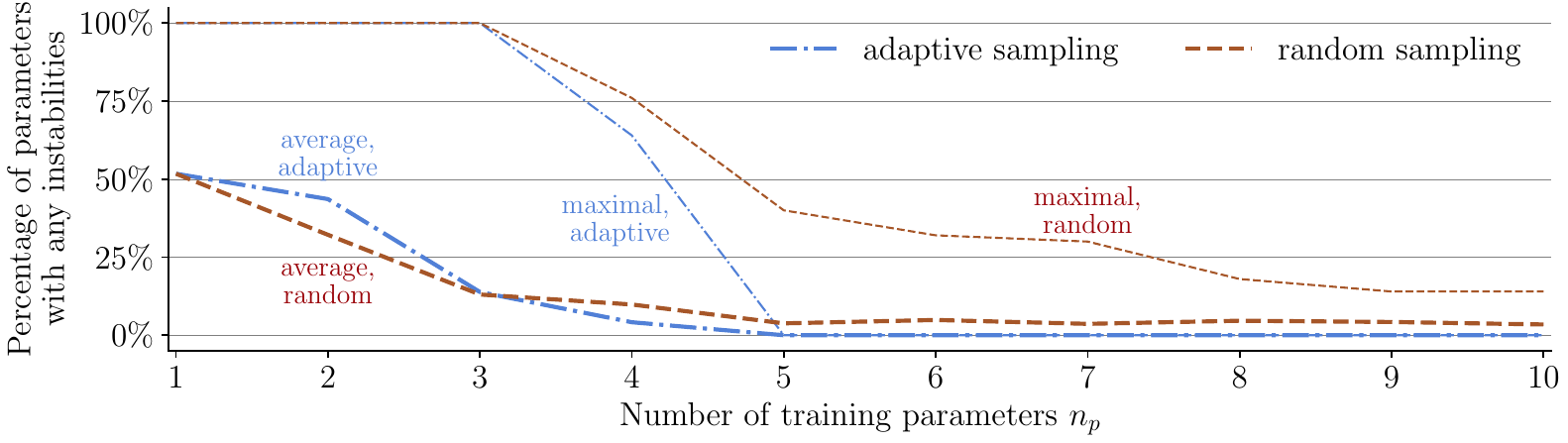}
    \caption{Percentages of the $50$ training parameter candidates where any instabilities are observed from $50$ posterior draws of the probabilistic ROM, as a function of the number of training parameters $n_p$, for the problem \cref{eq:burgers2D}. The results are averaged and maximized over $50$ trials with different random initial parameter samples.}
    \label{fig:burgers-instability}
\end{figure}

\begin{figure}[t]
    \centering
    \includegraphics[width=\textwidth]{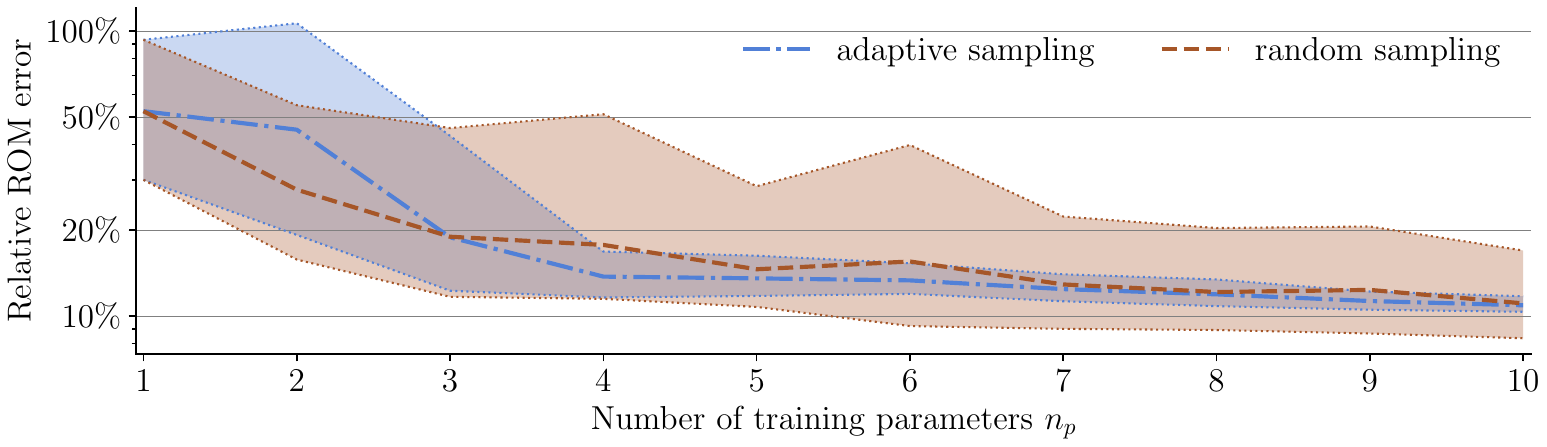}
    \caption{Total relative ROM errors $\mathcal{E}_\textrm{total}$ as a function of the number of training parameter samples, aggregated across 50 trials with different initial samples, for the problem \cref{eq:burgers2D}. The shaded regions show the interquantile range of $5\%$--$95\%$ of trials.
    }
    \label{fig:burgers-romerror}
\end{figure}

\begin{figure}[t]
    \centering
    \includegraphics[width=\textwidth]{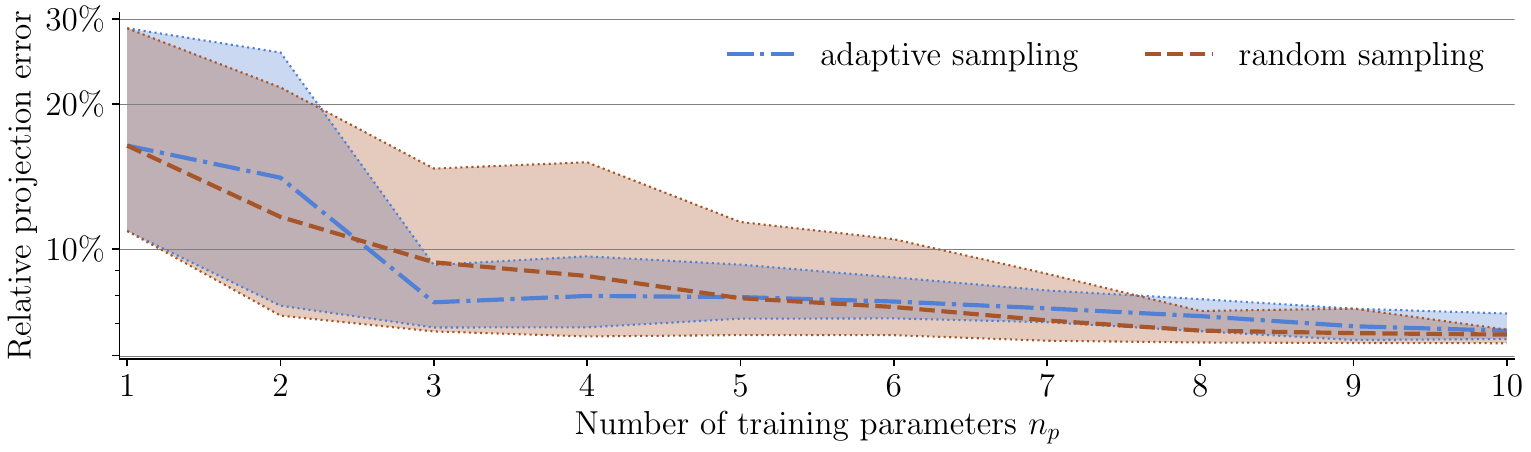}
    \caption{Total relative projection errors $\mathcal{E}_\textrm{proj}$ as a function of the number of training parameter samples, aggregated across 50 trials with different initial samples, for the problem \cref{eq:burgers2D}. The shaded regions show the interquantile range of $5\%$--$95\%$ of trials.}
    \label{fig:burgers-projectionerror}
\end{figure}

We repeat the procedure used to generate \Cref{fig:heat-instability} to analyze ROM stability in this problem, resulting in \Cref{fig:burgers-instability}. For $n_p = 2$, random sampling produces slightly more stable ROMs than adaptive sampling on average, likely because the second adaptively chosen training parameter sample is often the smallest or largest possible value for $\nu$, and hence as different as possible from the initial parameter sample. On the other hand, stability over the entire parameter domain is difficult to achieve with random sampling, but is achieved in every trial with $n_p = 5$ adaptively chosen training parameter samples. In terms of ROM error, shown in \Cref{fig:burgers-romerror}, the total error is slightly better on average with adaptive sampling than with random sampling for $n_p \ge 3$. The more important observation, however, is that adaptive sampling leads to accurate ROMs much more consistently than random sampling does. While some random sampling trials result in more accurate ROMs than the adaptive sampling ROMs, other random sampling trials also result in accuracy that is worse by about $10\%$ compared to the adaptive sampling ROMs. As in the previous experiment, the projection errors, shown in \Cref{fig:burgers-projectionerror}, are largely the same with both sampling strategies, but with adaptive sampling resulting in less variance across trials.

\section{Conclusion}\label{sec:conclusion}
We have introduced an uncertainty-aware active learning method for constructing data-driven parametric reduced-order models, tackling the challenge that ROM accuracy is sensitive to training data while full-order simulations are costly and hence the amount of available training data is limited. Our approach adaptively selects training parameters based on two proposed acquisition functions ensuring model stability and targeting regions of maximum predictive uncertainty. Numerical experiments on two nonlinear partial differential equations for thermal analysis and shock propagation demonstrate that our approach outperforms principled random sampling, achieving comparable ROM accuracy and superior stability with only about half as many full-order solves. Our approach shows significantly more robust results when repeated over different initial training sets, that is, the adaptive sampling scheme consistently leads to stable and accurate ROMs with as few samples as possible. While our applications focused on parameters appearing affinely in the operators, the methodology readily extends to parameters in initial conditions or input terms. Overall, we provide a general active learning framework for data-driven model reduction that leverages prediction uncertainty to produce efficient, robust, and uncertainty-informed reduced order models.

Future work may include exploring alternate acquisition functions for goal-oriented design, moving beyond greedy sequential selection to lookahead and batch selection strategies, and extending the framework to other ROM types beyond affine-structured operator inference. In particular, OpInf methods that guarantee stability, for example those in \cite{errico2026gasnitrom,goyal2025stablequadratic}, may improve our framework by allowing the adaptive sampling to focus on minimizing uncertainty rather than establishing stability. However, such an approach would require 1) extending these stability-focused methods to the parametric setting, and 2) constraining the posterior distribution from which the ROM is constructed to yield desired stability properties (for instance, enforcing symmetry or spectral properties in individual operators). Both of these goals remains a subject for future work. In general, however, the proposed active learning acquisition functions are directly applicable to any ROMs that provide prediction uncertainty estimates.

Another important challenge to address is working with higher-dimensional parameter spaces. Our proposed method is applicable to any set of parameter candidates (a strict grid layout is not required, see \Cref{remark:grids}), and sampling techniques such as sparse grid selection \cite{bungartz2004sparsegrids,farcas2018sparseUQ,peherstorfer2018multifidelitysurvey} can be used to construct a reasonable candidate set even in higher dimensions. However, constructing parametric ROMs that remain accurate over high-dimensional parameter domains, without resorting to domain decomposition or interpolation-based ROMs, remains an open challenge \cite{benner2015pmorsurvey,mcquarrie2023parametric}. Approaches for reducing the dimensionality of the parameter space may be paired with our framework since no geometric structure is required in the set of candidates. A related bottleneck for POD-based ROMs is that they can only be as good as the dimensionality reduction driven by linear manifolds, a problem that is exacerbated when the system behavior changes significantly with the parameters. Incorporating more sophisticated dimensionality reduction through nonlinear manifolds~\cite{barnett2022quadmanifold,geelen2023quadmanifold,jain2017quadratic,schwerdtner2024greedy}, time- or parameter-adaptive bases \cite{amsallem2008grassmannianrom,dechant2026dynamicsubspaces,dosSantos2022grassmanniandiffusion}, or by jointly optimizing the bases with ROM operators \cite{errico2026gasnitrom,padovan2024nitrom,schwerdtner2025opinfawarequadmanifold} are all potential directions to further improve the framework, especially for problems featuring higher-dimensional parameters.

\section*{Acknowledgments}
A.C. acknowledges support from AFOSR grant FA9550-24-1-0327 under the Multidisciplinary University Research Initiatives (MURI), the NSF FDT-Biotech award 2436499, and the Texas Institute for Electronics (TIE) DARPA \#HR00112430347.

\bibliographystyle{siamplain}
\bibliography{references}
\end{document}